%% file: signal2image_traits.tex
\pdfoutput=1

\documentclass[preprint,12pt]{elsarticle}

\usepackage[a4paper, margin=1in]{geometry}
\usepackage{graphicx}
\graphicspath{{figures/}}
\usepackage{amsmath}
\usepackage{hyperref}
\usepackage{longtable}
\usepackage{booktabs}
\usepackage{caption}
\usepackage{subcaption}
\usepackage{overpic}
\usepackage[utf8]{inputenc}
\usepackage[T1]{fontenc}
\usepackage{textcomp}

\usepackage{xcolor}

\newcommand{\jl}[1]{#1}

\usepackage{lineno}
\leftlinenumbers

\usepackage{array}
\usepackage{enumitem}
\usepackage{float}

\usepackage{textcomp}
\usepackage{gensymb}

\biboptions{authoryear,round,semicolon}

\title{Turning spectra into images improves plant trait retrieval with 2D-CNNs}

\author[1,2,3]{Javier Lopatin\corref{cor1}}
\ead{javier.lopatin@uai.cl}

\author[4]{Teja Kattenborn}
\ead{teja.kattenborn@geosense.uni-freiburg.de}

\author[5]{Eya Cherif}
\ead{eya.cherif14@gmail.com}

\author[1]{Sebastián Moreno}
\ead{sebastian.moreno@uai.cl}

\address[1]{Faculty of Engineering and Science, Adolfo Ibáñez University, Av. Diagonal Las Torres 2460, Santiago, Chile, 7941169}

\address[2]{Data Observatory Foundation, ANID Technology Center No. DO210001, Santiago, Chile, 7510277}

\address[3]{Center for Climate Resilience Research (CR)2, University of Chile, Santiago, Chile, 8370449}

\address[4]{Sensor-based Geoinformatics (geosense), University of Freiburg, Freiburg, Breisgau, Germany}

\address[5]{Institute for Earth System Science and Remote Sensing, Leipzig University, Germany}

\cortext[cor1]{Corresponding author; javier.lopatin@uai.cl}

\begin{document}
\begin{frontmatter}

\begin{abstract}
Hyperspectral reflectance spectroscopy enables non-destructive estimation of plant functional traits, yet current deep learning approaches process spectra as one-dimensional sequences, which limits their ability to capture long-range inter-band dependencies. We asked whether transforming 1D spectra into 2D image representations improves multi-trait prediction with convolutional neural networks (CNN). We compared nine transformations using EfficientNet-B0 on the GreenHyperSpectra dataset (7,897 labeled spectra, eight traits, 400--2450 nm), benchmarked against published 1D CNN results on the same split. Trained from scratch, the simplest transformation, a direct Reshape of the spectrum into a 2D grid, performed best ($R^2 = 0.684 \pm 0.001$) and improved on the state-of-the-art 1D baseline ($R^2 = 0.587$, $+0.097$). We then pretrained a 2D masked autoencoder (MAE-2D) on 139,000 unlabeled spectral images. Linear probing, which freezes the encoder and trains only a multilayer perceptron head, reached $R^2 = 0.646$ and exceeded every 1D self-supervised counterpart, including the fine-tuned MAE-1D ($R^2 = 0.641$). We did not find significant differences compared to the 1D baselines in cross-dataset evaluation. To identify which wavelengths drive each prediction, we applied Integrated Gradients and Grad-CAM and unfolded band importance back to the spectral axis. Protein ($r = 0.45$) and leaf water ($r = 0.33$) agreed with sensitivities simulated by the PROSAIL radiative-transfer model, while carotenoids ($r = 0.06$) and leaf area index ($r = -0.11$) did not, showing that the model reads established leaf chemistry for traits with sharp absorption features. The representational advantage of 2D spectral images, rather than architectural complexity or ImageNet pretraining, drives the gain over 1D approaches.

\end{abstract}


\end{frontmatter}


\section{Introduction}
\label{sec:introduction}

Plant functional traits are measurable properties that influence how organisms respond to and affect their environment, and are central to understanding ecosystem functioning, biodiversity, and the global carbon cycle \citep{kattge2020try}. In particular, traits such as leaf chlorophyll content, carotenoid concentration, leaf mass per area, and leaf area index govern photosynthetic capacity, light interception, and nutrient cycling, making their accurate quantification essential for ecological research and precision agriculture \citep{homolova2013review}. To meet this need, hyperspectral reflectance spectroscopy offers a non-destructive pathway for estimating these traits at scale, since foliar biochemistry and canopy structure imprint characteristic absorption features across the 400--2500\,nm spectral range \citep{curran1989imaging, fu2020photosynthetic}. 

Traditional approaches for deriving traits from spectra rely on vegetation indices, look-up table inversion of radiative transfer models such as PROSAIL \citep{feret2021prospectpro, ferreira2025enhancing}, or statistical methods, principally partial least squares regression (PLSR), that project the high-dimensional spectral space onto a low-dimensional latent representation \citep{fu2020photosynthetic, ji2024unveiling}. While effective, PLSR models are typically trained and validated on individual traits, require careful feature engineering, and can struggle to capture the non-linear relationships between reflectance and biochemical concentrations that arise from multiple scattering and overlapping absorption features \citep{verhoef2007sail}. More recently, one-dimensional convolutional neural networks (1D-CNNs) have been applied directly to spectral vectors, learning hierarchical features end-to-end and achieving state-of-the-art multi-trait retrieval performance without manual feature selection \citep{cherif2023spectra, furbank2021wheat}. For example, \citet{cherif2023spectra} demonstrated that a single EfficientNet-based 1D-CNN trained simultaneously on 20 traits across 42 heterogeneous field datasets outperformed PLSR baselines, establishing a benchmark for transferable multi-trait models. Their subsequent \texttt{GreenHyperSpectra} dataset \citep{cherif2025greenhyperspectra} further advanced this line of work by incorporating semi-supervised and self-supervised pretraining strategies, with a 1D masked autoencoder (MAE) achieving the best results ($R^2 = 0.641$) after fine-tuning on labeled data.

Despite this progress, 1D-CNNs process the spectrum as a flat sequence, inherently limiting their receptive field to local neighborhoods of adjacent bands. Spectral signatures of plant traits, however, arise from the interplay of multiple absorption features distributed across the full wavelength range, for example, the joint influence of pigment absorption in the visible, water absorption in the shortwave infrared, and structural scattering near the red edge \citep{curran1989imaging, homolova2013review}. Capturing these long-range inter-band dependencies within a 1D framework requires either very deep networks or large kernel sizes, both of which increase the risk of overfitting when labeled training data are scarce.

\begin{figure}[!tbp]
    \centering
    \includegraphics[width=\textwidth]{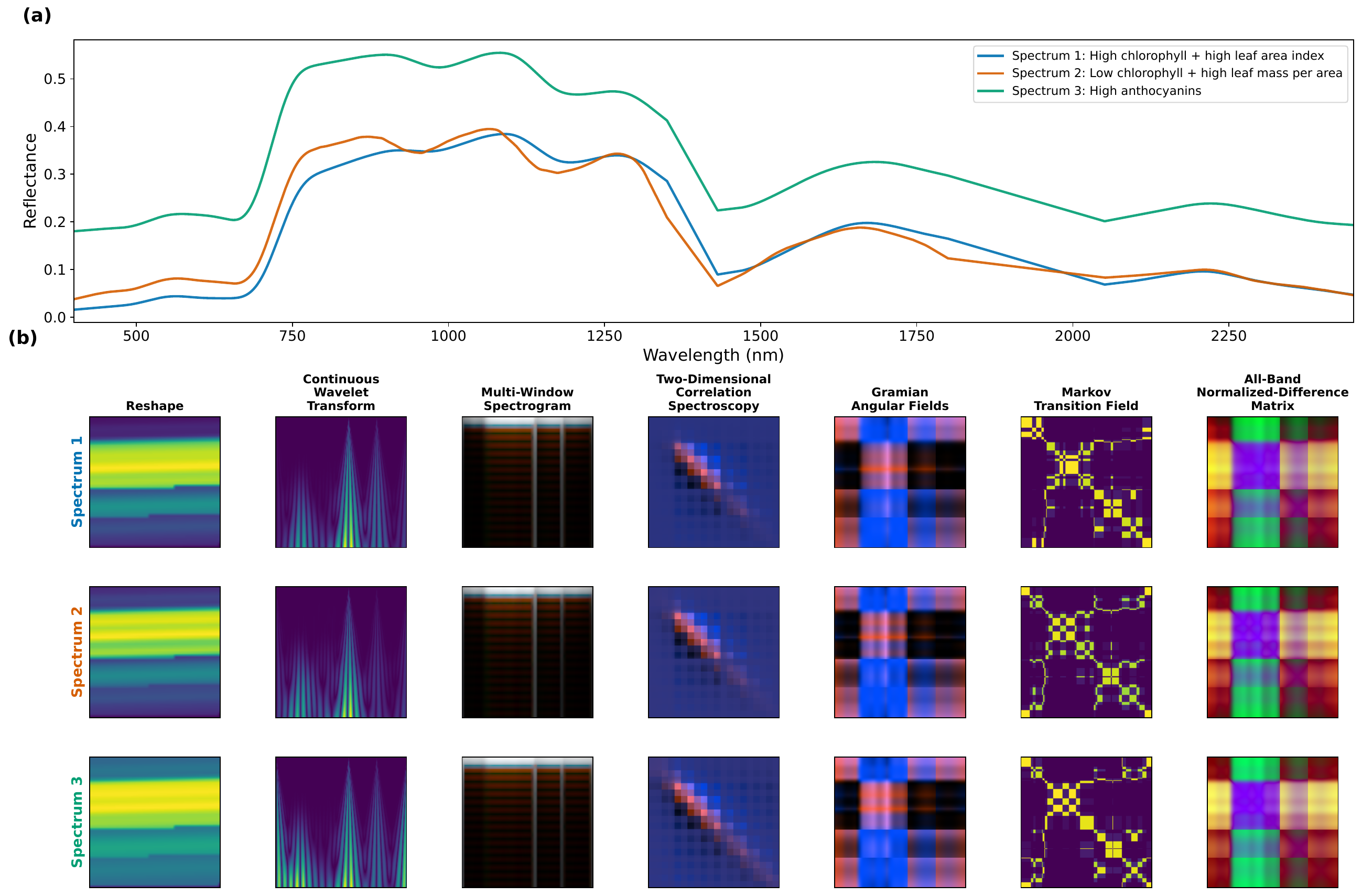}
    \caption{Illustration of seven of the nine 1D-to-2D spectral transformations applied to three representative spectra with contrasting trait profiles. The Serpentine and Hilbert-curve variants are re-layouts of the same Reshape values and are not shown separately. (a) Original reflectance spectra: Spectrum~1 exhibits high chlorophyll content and high leaf area index (LAI); strong visible absorption and a near-infrared plateau. Spectrum~2 shows low chlorophyll and high leaf mass per area (reduced visible absorption, prominent dry matter features), and Spectrum~3 displays high anthocyanin content (elevated green-edge reflectance). (b) Corresponding 2D image representations produced by each transformation method. Each transformation encodes different aspects of the spectral information: Reshape preserves wavelength adjacency as spatial texture; Continuous wavelet transform (CWT) reveals multi-scale absorption features; Spectrogram captures localized frequency content through three complementary windows; 2D-COS highlights inter-band correlations; Gramian Angular Fields (GAF) encodes angular relationships between spectral values; Markov Transition Fields (MTF) represent transition dynamics between discretized intensity bins; and the all-band normalized-difference matrix (NDI) encodes every pair of bands as a normalized-difference index. Reflectance is not available between 1351--1430 and 1801--2050 nm, where the water
    absorption regions were excluded from the dataset. The curves are drawn continuously across those two intervals, so the straight segments there are interpolation, not
measured signal.}.
    \label{fig:transforms_example}
\end{figure}

An alternative strategy, increasingly explored in signal processing and chemometrics, is to transform one-dimensional signals into two-dimensional image representations and leverage the powerful feature extraction capabilities of 2D-CNNs originally developed for computer vision \citep{wang2015gaf, garcia2022temporal, Tan2026-vx}. Several encoding methods have been proposed for this purpose. Gramian Angular Fields (GAF) and Markov Transition Fields (MTF) encode temporal or spectral correlations as square images by mapping signal values to angular or transition-probability spaces \citep{wang2015gaf, ganesan2021gaf}. Continuous wavelet transform (CWT) scalograms decompose signals across multiple frequency scales, producing time-frequency representations that have proven effective for fault diagnosis, medical signal classification, and spectral analysis \citep{addison2017wavelet, Lopatin2023-fn, ong2025sugarcane, zhuang2023cwt}. Short-time Fourier transform spectrograms offer a complementary multi-window decomposition \citep{shuai2025multichannel}, while direct reshaping, simply folding the 1D vector into a 2D matrix, has been shown to produce surprisingly competitive results by preserving wavelength adjacency as spatial texture \citep{hennessy2022reshaping, deev2024spectrum, sharma2019deepinsight}. Building on this idea, band adjacency can be preserved more faithfully with alternating raster (serpentine) or space-filling (Hilbert-curve) layouts \citep{kamata1995hilbert}, whereas all-band normalized-difference matrices instead encode every pair of bands, extending the logic of two-band spectral indices \citep{inoue2008ndsi}. More recently, these encoding techniques have begun to appear in spectroscopic applications, including GAF-based adulteration detection in food products \citep{zhang2025gaf} and CWT-assisted disease recognition from near-infrared spectra \citep{ong2025sugarcane}. Figure \ref{fig:transforms_example} shows seven transformation examples for three different vegetation signals. The central premise is that converting one-dimensional hyperspectral reflectance signals into two-dimensional image representations may reveal inter-band relationships, multi-scale spectral patterns, and structural regularities that conventional 1D convolutional architectures fail to capture.

However, the application of 1D-to-2D spectral transformations to vegetation trait retrieval from hyperspectral data remains largely unexplored. Existing studies in this domain have predominantly focused on classification tasks (e.g., species discrimination) rather than continuous regression of biophysical variables \citep{hennessy2022reshaping}, and no systematic comparison of multiple transformation methods under controlled experimental conditions has been conducted. 
Furthermore, converting spectra into images enables the direct application of established 2D vision architectures and self-supervised learning frameworks. Among these, 2D masked autoencoders \citep[i.e., MAEs;][]{he2022mae} have demonstrated strong representation-learning capabilities using unlabeled image data. Although analogous 1D MAEs provide an effective alternative for spectral pretraining \citep{cherif2025greenhyperspectra}, it remains unknown whether applying 2D MAEs to transformed spectral images offers representational advantages for vegetation trait retrieval.

In this study, we aim to: (i) systematically compare the performance of nine 1D-to-2D spectral transformation methods as spectral encoding strategies for 2D-CNN-based trait retrieval; (ii) assess whether converting spectra to 2D images enables self-supervised pretraining via masked autoencoders to improve trait predictions beyond what 1D pretraining achieves; (iii) assess whether the 2D representation improves the transferability to unseen sites; and (iv) create a variable importance visualization from the 2D representations and test it against radiative-transfer sensitivity. We used the \texttt{GreenHyperSpectra} dataset to benchmark our 2D approach against the state-of-the-art 1D methods \citep{cherif2025greenhyperspectra}. 

\section{Methods}
\label{sec:methods}

\subsection{Study Data}
\label{sec:data}

We used the \texttt{GreenHyperSpectra} dataset \citep{cherif2025greenhyperspectra}, a large-scale multi-source hyperspectral collection recently published as a benchmarking resource for plant trait prediction. The dataset comprises two components: a labeled set of 7,897 vegetation canopy reflectance spectra with co-located measurements of seven functional plant traits, and an unlabeled set of 139,295 spectra collected from proximal, airborne, and spaceborne platforms across diverse continents, ecosystems, and sensor configurations. All spectra were resampled to a uniform wavelength grid spanning 400--2450\,nm at 1\,nm resolution. Regions of strong atmospheric water absorption (1351--1430\,nm, 1801--2050\,nm, and 2451--2500\,nm) were removed, and a Savitzky-Golay smoothing filter (window = 65, polynomial order = 1) was applied per spectral segment, yielding 1721 spectral bands per sample \citep{cherif2025greenhyperspectra}. Figure \ref{fig:transforms_example} a) shows an example of three input data points/spectra. As can be observed, the wavelength ranges from 400 to 2450\,nm and, for each wavelength, we show its corresponding reflectance value (the input data).

The labeled subset aggregates data from 50 field campaigns spanning forests, grasslands, croplands, tundra, and pastures, with trait values harmonized to area-based units. The seven target traits, outputs to predict, correspond to parameters of the PROSAIL-PRO radiative transfer model \citep{feret2021prospectpro}: leaf chlorophyll content (C\textsubscript{ab}, $\mu$g\,cm\textsuperscript{$-$2}), carotenoid content (C\textsubscript{ar}, $\mu$g\,cm\textsuperscript{$-$2}), anthocyanin content (C\textsubscript{anth}, $\mu$g\,cm\textsuperscript{$-$2}), equivalent water thickness (C\textsubscript{w}, g\,m\textsuperscript{$-$2}), leaf mass per area (C\textsubscript{m}, g\,m\textsuperscript{$-$2}), leaf area index (LAI, m\textsuperscript{2}\,m\textsuperscript{$-$2}), and protein content (C\textsubscript{p}, g\,m\textsuperscript{$-$2}). An eighth derived trait, carbon-based constituents (C\textsubscript{bc} = C\textsubscript{m} $-$ C\textsubscript{p}), was additionally computed following \citet{cherif2025greenhyperspectra}. Not all samples possess measurements for every trait; the dataset is inherently sparse, with missing values handled through masked loss functions during training (Section~\ref{sec:training}).

\subsection{1D-to-2D Spectral Transformations}
\label{sec:transforms}

We evaluated nine transformation methods, each encoding different aspects of the spectral information into spatial image structure. All transformations produced images of $224 \times 224$ pixels (with varying numbers of channels) suitable for standard convolutional neural network architectures (Figure~\ref{fig:transforms_example}).

\subsubsection{Direct Reshape}

The simplest transformation reshapes the 1D spectral vector into a 2D matrix by computing the smallest integer side length $s = \lceil\sqrt{n}\rceil$ where $n = 1721$ bands, yielding $s = 42$. The spectrum is zero-padded to $42 \times 42 = 1764$ elements (appending 43 zeros) and folded into a square matrix, which is then bilinearly resized to $224 \times 224$ and min-max normalized to $[0, 1]$. The scaling uses each image's own minimum and maximum rather than statistics pooled over the training set, so the transform preserves spectral shape and discards overall amplitude. We also tested a global scaling fitted on the training set, which lowered accuracy for all eight traits (Table~S7), so we kept the per-image version. Despite its simplicity, this transformation preserves the sequential ordering of wavelengths, such that spatially adjacent pixels correspond to spectrally proximate bands \citep{hennessy2022reshaping}. The resulting single-channel image encodes local spectral gradients as spatial texture that 2D convolutions can exploit.

\subsubsection{Serpentine and Hilbert-Curve Reshape}

The direct Reshape preserves the global wavelength order, but its row-major raster breaks spectral adjacency at every row boundary, because the last band of one row and the first band of the next fall far apart in the image. We tested two reshape variants \citep{hennessy2022reshaping} that remove this discontinuity. The Serpentine, or boustrophedon, reshape lays out consecutive rows in alternating direction, reversing every odd row, so that consecutive bands always map to spatially adjacent pixels, including across rows. The Hilbert-curve reshape instead places the bands along a Hilbert space-filling curve, which keeps points that are close along the spectrum close in the plane at every scale \citep{kamata1995hilbert}. Because the Hilbert curve needs a power-of-two side, we resampled the spectrum to $64 \times 64 = 4096$ points by linear interpolation and filled the grid completely, avoiding a large zero-padded region. Both variants produce a single-channel $224 \times 224$ image, normalized as for the direct Reshape.

\subsubsection{Continuous Wavelet Transform}

The continuous wavelet transform (CWT) decomposes the spectral signal across multiple frequency scales simultaneously, producing a time-frequency (or in this case, wavelength-scale) representation known as a scalogram \citep{addison2017wavelet}. We employed the Morlet wavelet across 128 logarithmically spaced scales ranging from 1 to 128, computed using the PyWavelets library \citep{pywt}. The absolute values of the wavelet coefficients form a $128 \times 1721$ scalogram that captures both narrow absorption features (at fine scales) and broad spectral trends (at coarse scales). The scalogram was bilinearly resized to $224 \times 224$ and min-max normalized, yielding a single-channel image. CWT approaches have proven effective in capturing multi-scale spectral features for trait retrieval at the canopy \citep{Blackburn2007-vn} and leaf \citep{Lopatin2023-fn} levels.

\subsubsection{Multi-Window Spectrogram}

The short-time Fourier transform (STFT) was applied with three different window functions---Bartlett, Gaussian ($\sigma = 12$), and Blackman---each producing a complementary time-frequency representation of the spectrum. We used a segment length of 64 samples with 50\% overlap (32 samples), following the multi-channel approach of \citet{shuai2025multichannel}. The magnitude of each STFT output was computed, bilinearly resized to $224 \times 224$, and min-max normalized. The three resulting spectrograms were stacked as a three-channel (RGB-like) image, where each channel emphasizes different spectral features owing to the distinct frequency leakage and sidelobe characteristics of the respective windows.

\subsubsection{Two-Dimensional Correlation Spectroscopy}

Two-dimensional correlation spectroscopy (2D-COS) generates synchronous and asynchronous correlation maps from perturbation-induced spectral variations \citep{noda2004twodimensional}. In classical 2D-COS, perturbation arises from external stimuli (e.g., temperature, concentration) applied across multiple spectra. Here, we adapted the formulation for single-sample analysis by simulating perturbation through dividing each spectrum into 10 overlapping segments (50\% overlap) and computing the cross-correlation between band intensities across segments. The synchronous correlation matrix $\Phi = \mathbf{X}^T \mathbf{X} / (m-1)$, where $\mathbf{X}$ is the mean-centered segment matrix and $m$ is the number of segments, captures in-phase inter-band correlations. The asynchronous correlation matrix $\Psi = \mathbf{X}^T \mathbf{N} \mathbf{X} / (m-1)$ employs the Hilbert-Noda transformation matrix $N_{ij} = 1/[\pi(j-i)]$ for $i \neq j$ (and zero otherwise), with $i$ and $j$ indexing segments, revealing sequential spectral changes. Both matrices were individually min-max normalized and bilinearly resized to $224 \times 224$, then stacked as a two-channel image.

\subsubsection{Gramian Angular Fields}

Gramian Angular Fields (GAF) encode temporal (or spectral) correlations as angular relationships in a polar coordinate system \citep{wang2015gaf}. The spectrum is first downsampled to 224 values via linear interpolation and scaled to $[-1, 1]$. An angular representation $\varphi_i = \arccos(x_i)$ is computed for each spectral value, and two complementary fields are generated: the Gramian Angular Summation Field (GASF), defined as $\text{GASF}_{ij} = \cos(\varphi_i + \varphi_j)$, which captures in-phase spectral correlations; and the Gramian Angular Difference Field (GADF), defined as $\text{GADF}_{ij} = \sin(\varphi_i - \varphi_j)$, which captures phase differences. The two fields were min-max normalized and stacked as a two-channel image.

\subsubsection{Markov Transition Field}

The Markov Transition Field (MTF) encodes the transition dynamics between discretized spectral value bins as a spatial image \citep{wang2015gaf}. The spectrum is downsampled to 224 values and discretized into 8 quantile-based bins. A first-order Markov transition matrix $\mathbf{P}$ is estimated, where $P_{ij}$ represents the conditional probability of transitioning from bin $i$ to bin $j$ across consecutive spectral positions. The MTF is then constructed as $\text{MTF}_{ij} = P[q_i, q_j]$, where $q_i$ and $q_j$ are the bin assignments at spectral positions $i$ and $j$, respectively. The resulting $224 \times 224$ single-channel image was min-max normalized.

\subsubsection{All-Band Normalized-Difference Matrix}

The all-band normalized-difference matrix (NDI) tests the inter-band interaction hypothesis directly, because every pixel is an explicit function of two arbitrarily distant bands. Following the logic of two-band normalized-difference indices, which have been optimized over all band pairs for trait retrieval \citep{inoue2008ndsi}, each pixel $(i, j)$ encodes the relationship between bands $\lambda_i$ and $\lambda_j$. We first downsampled the spectrum to 224 bands by uniform index sampling to keep the all-pairs matrix tractable. Because the normalized difference discards absolute reflectance, we stacked three complementary channels: the normalized difference $(R_i - R_j)/(R_i + R_j + \varepsilon)$, the absolute difference $|R_i - R_j|$, and the geometric level $\sqrt{R_i R_j}$, with $\varepsilon = 10^{-8}$. Each channel was min-max normalized to $[0, 1]$ independently, giving a three-channel $224 \times 224$ image.

\subsubsection{Composite Input}

We used a composite approach to test whether combining complementary encodings recovers information that any single transform misses, stacking several transforms as separate image channels so one network sees them at once. Because the transforms encode overlapping information and yield highly correlated predictions, we let complementarity rather than individual accuracy guide the choice of channels. We grouped the transforms into three information families, a raw spatial-layout family (Reshape, Serpentine, Hilbert), a time-frequency family (CWT and spectrogram), and a pairwise inter-band family (NDI, GAF, and 2D-COS), and built a composite from the best-performing member of each family, as identified by the single-transform comparison (Section~\ref{sec:results_transforms}). 

\subsection{Model Architectures}
\label{sec:models}
Having converted each spectrum into a 2D image, we now apply image models to predict the traits. We treated the task as a single multi-output regression, where one model predicts all eight traits (Section~\ref{sec:data}) at once instead of training eight separate models. We tested two settings. We first trained a supervised CNN from scratch. Then, we pretrained a self-supervised MAE on unlabeled images before fine-tuning it on the labeled data.

\subsubsection{Supervised Regression: EfficientNet-B0}
\label{sec:efficientnet}

For supervised multi-trait regression, we employed EfficientNet-B0 \citep{tan2019efficientnet} as the backbone architecture, instantiated through the timm library \citep{rw2019timm} and trained from scratch. We selected EfficientNet-B0 for its favorable trade-off between model capacity ($\sim$4\,M trainable parameters) and computational efficiency, which proved critical given the moderate size of the labeled training set ($n = 4{,}508$). It is also the selected model for the 1D baseline in \citet{cherif2025greenhyperspectra}. 
 We replaced the network's final classification layer with a single linear head mapping to 8 output neurons (one per target trait), so one network predicts all traits jointly, and applied a dropout rate of 0.3 before the output layer. For single-channel transformations (e.g., Reshape, CWT, MTF), we adapted the first convolutional layer from three to one input channel by averaging the randomly initialized channel weights along the input dimension. Multi-channel transformations (e.g., Spectrogram, 2D-COS, GAF) required analogous channel-count adjustments.

\subsubsection{Self-Supervised Pretraining: Masked Autoencoder (MAE-2D)}
\label{sec:mae}

To leverage the 139,295 unlabeled spectra for representation learning, we implemented a two-dimensional Masked Autoencoder (MAE) following the framework of \citet{he2022mae}. The MAE operates on 2D spectral images produced by the best-performing transformation (Reshape; see Section~\ref{sec:results_transforms}), enabling the use of standard vision transformer components.

The encoder follows a Vision Transformer (ViT) architecture \citep{dosovitskiy2021vit} with patch size $16 \times 16$, embedding dimension 192, 6 transformer blocks, and 3 attention heads per block, yielding 196 patches per $224 \times 224$ image. During pretraining, 75\% of patches are randomly masked, and only the visible 25\% are processed by the encoder, substantially reducing computational cost. Fixed sinusoidal 2D positional embeddings are added to all patch tokens, and a learnable class (CLS) token is prepended to the sequence. The CLS token plays no classification role here. It acts as a learned pooling operator that attends to every visible patch, and its output embedding serves as the global representation of the image, an alternative to averaging the patch tokens.

The decoder is a lightweight transformer with embedding dimension 96, 4 blocks, and 3 attention heads. It receives the encoded visible patches along with learnable mask tokens that replace the masked positions. After processing through the decoder blocks, a linear projection maps each token to the pixel values of its corresponding $16 \times 16$ patch. The reconstruction loss is computed only on masked patches:
\begin{equation}
    \mathcal{L}_\text{MAE} = \frac{1}{|\mathcal{M}|} \sum_{i \in \mathcal{M}} \| \hat{\mathbf{p}}_i - \mathbf{p}_i \|^2
    \label{eq:mae_loss}
\end{equation}
where $\mathcal{M}$ denotes the set of masked patch indices, $\hat{\mathbf{p}}_i$ is the reconstructed patch, and $\mathbf{p}_i$ is the original patch. We did not apply any data augmentation beyond random masking during pretraining
, consistent with the finding of \citet{he2022mae} that masking itself provides sufficient regularization, and following the protocol of \citet{cherif2025greenhyperspectra} whose 1D MAE likewise omitted augmentation.

The total model comprises approximately 3.3\,M parameters (2.8\,M encoder, 0.5\,M decoder). After pretraining, the decoder is discarded, and the encoder's token representation serves as a global feature vector for downstream trait regression through an appended two-layer MLP head (192 $\rightarrow$ 192 $\rightarrow$ 8, with GELU activation and 10\% dropout).

\subsection{Training Protocol}
\label{sec:training}

For model training and evaluation, we adopted the standardized data split released with GreenHyperSpectra, which partitions 5,635 of the labeled spectra with an 80/20 hold-out into 4,508 training and 1,127 test samples, with stratification by source dataset to maintain proportional representation of vegetation types, sensors, and acquisition conditions across both subsets. This is the same fixed split, with identical training and test membership, that \citet{cherif2025greenhyperspectra} used to report their 1D baselines, so our 2D models and their 1D results are trained and evaluated on exactly the same data. To quantify variability due to stochastic training effects, we trained all models under three random seeds, following the evaluation protocol of \citet{cherif2025greenhyperspectra}.

We implemented all models in PyTorch 2.6 \citep{paszke2019pytorch} using PyTorch Lightning 2.6 for training orchestration, and stored the pre-computed 2D transformations as memory-mapped arrays to avoid recomputation during training.

\subsubsection{Supervised Training}

We optimized the supervised EfficientNet-B0 models with AdamW \citep{loshchilov2019adamw} ($\beta_1 = 0.9$, $\beta_2 = 0.999$, weight decay = $10^{-4}$) under a cosine annealing learning rate schedule with initial rate $10^{-3}$ and minimum rate $10^{-6}$. Training proceeded for a maximum of 100 epochs with early stopping (patience = 15 epochs, monitoring validation loss). Within the training set, we withheld 15\% of samples for validation by random splitting and set the batch size to 16.

Because the trait label matrix is sparse---not all samples have measurements for all eight traits---we employed a masked mean squared error (MSE) loss that computes gradients only over observed trait values:
\begin{equation}
    \mathcal{L}_\text{masked} = \frac{\sum_{i,j} m_{ij}(\hat{y}_{ij} - y_{ij})^2}{\sum_{i,j} m_{ij}}
    \label{eq:masked_mse}
\end{equation}
where $\hat{y}_{ij}$ and $y_{ij}$ are the predicted and observed values for sample $i$ and trait $j$, and $m_{ij}$ is a binary mask indicating whether the observation is available. This mask flags missing trait labels and has no relation to the random patch masking of Equation~\ref{eq:mae_loss}, which hides parts of the input image. Similar to \citet{cherif2023spectra}, during training, we applied two forms of data augmentation to the training set: (i) an additive baseline shift drawn uniformly from $\pm$2\% of the spectral range, and (ii) a multiplicative scaling factor drawn uniformly from $[0.98, 1.02]$, both applied to the raw spectra before transformation. These perturbations are small and label-preserving, because they emulate illumination and sensor variation that leaves the underlying trait unchanged, so the trait label stays valid. 
For pre-computed (cached) images, we added small Gaussian noise ($\sigma = 0.01$) to the transformed images at each epoch. The transform comparison used the cached images, so every transform received the same image-space augmentation (Gaussian noise, $\sigma = 0.01$), while the spectrum-space baseline and multiplicative perturbations applied only to the on-the-fly path. Augmentation was therefore identical across the compared transforms.Prior to training, we transformed trait values with the Yeo-Johnson power transformation \citep{yeo2000power} fitted on the training set, which stabilizes variance and reduces the influence of skewed distributions. We inverse-transformed the predictions for evaluation in original units. 

\subsubsection{MAE Pretraining}

We pretrained the MAE-2D on Reshape images derived from the 139,295 unlabeled spectra (after filtering 375 samples with missing data, yielding 138,920 valid images). Training used AdamW ($lr = 10^{-4}$, weight decay = $10^{-4}$) with cosine annealing over 300 epochs, a batch size of 32, and gradient clipping (max norm = 1.0). We used a 95/5 train/validation split to monitor reconstruction loss, with early stopping at patience = 30 epochs. The pretraining objective was pure MSE reconstruction of masked patches (Equation~\ref{eq:mae_loss}), without additional augmentation, as random masking already provides effective regularization \citep{he2022mae}.

\subsubsection{Fine-Tuning}

After MAE pretraining, we discarded the decoder and equipped the encoder with a regression MLP head. We fine-tuned the full model, updating all encoder parameters jointly with the regression head ($lr = 10^{-4}$). Fine-tuning used the same protocol as supervised training (AdamW, early stopping, masked MSE loss) with 100 maximum epochs. The head maps the 192-dimensional CLS embedding through a 192 $\rightarrow$ 192 $\rightarrow$ 8 multilayer perceptron with GELU activation and dropout 0.1, whose
eight outputs are the eight traits.

\subsection{Experimental Design}
\label{sec:experimental_design}

The experiments are organized into three complementary case studies that progressively demonstrate the advantages of 2D spectral representations.

\textit{Case Study 1: Transform comparison.} We evaluated all nine 1D-to-2D transformations (Section~\ref{sec:transforms}) with EfficientNet-B0 under identical supervised training conditions. The primary comparison is between our 2D approach and the supervised EfficientNet-1D baseline reported by \citet{cherif2025greenhyperspectra} (mean $R^2 = 0.587$ across eight traits). 

\textit{Case Study 2: Self-supervised pretraining.} Using the best transform methods, we assessed the impact of pretraining strategies. We used: (i) supervised training without pretraining (the Case Study 1 baseline); and (ii) MAE-2D pretraining on 139K unlabeled images followed by fine-tuning on real labeled data, compared against the MAE-1D of \citet{cherif2025greenhyperspectra} ($R^2 = 0.641$).

\textit{Case Study 3: Out-of-distribution generalization.} We used the cross-dataset protocol of \citet{cherif2025greenhyperspectra} to test whether our approach generalizes to unseen field campaigns. We held out five labeled sub-datasets at a time from the pool of 50 and trained on the rest with an internal 80/20 training and validation split, repeating over the 12 folds in which two datasets recur so that every one of the 50 is held out at least once. Because this used a single training run, the reported mean and standard deviation of $R^2$ come from their five balanced subsamples rather than from repeated training. We compared our best transformation model and our MAE-2D encoder against their matching 1D regimes, the supervised 1D-CNN and the fine-tuned MAE-1D, respectively.

\subsection{Evaluation Metrics}
\label{sec:metrics}

We evaluated model performance with the coefficient of determination ($R^2$) and the normalized root mean square error (nRMSE, expressed as a percentage of the 1st--99th percentile range of observed values), computed per trait on the held-out test set. We report the mean and standard deviation across the three random seeds. We averaged per-trait $R^2$ values to obtain a single summary statistic for each method, enabling direct comparison with the baselines of \citet{cherif2025greenhyperspectra}. Per-trait and per-method nRMSE, using this same normalization, are reported in the supplementary material (Tables~S1 to~S4). The remaining metrics, root mean square error, mean absolute error, and bias, are available in the code repository.

To assess significance, we compared each transform against the reported 1D baseline of \citet{cherif2025greenhyperspectra} using a Wilcoxon signed-rank test on the per-trait $R^2$ differences, paired across the eight traits. We chose a non-parametric paired test over a $t$-test because the eight traits are predicted jointly and are not independent, and we test against the published baseline rather than between our own transforms.

\subsection{Spectral Importance}
\label{sec:importance}

We used an attribution approach to obtain the variable importance of the predicted traits. We specifically applied two standard interpretability methods to the best supervised model (Reshape + EfficientNet-B0), Integrated Gradients (IG) \citep{Sundararajan2017-lv} and Grad-CAM \citep{Selvaraju2016-ba}. IG produces signed, input-level attributions, which we computed using the training-set mean image as the baseline and a 50-step Riemann sum. We attributed on the normalized image that the network receives rather than on the raw spectrum. The Reshape transform applies a per-image min-max scaling, so differentiating through that step piles the gradient onto the few bands that set the scale. Attributing on the network input avoids this trait-agnostic artifact and preserves the IG completeness axiom. Grad-CAM produces region-level relevance over the final convolutional embedding.

\begin{figure}[!htbp]
    \centering
    \includegraphics[width=\textwidth]{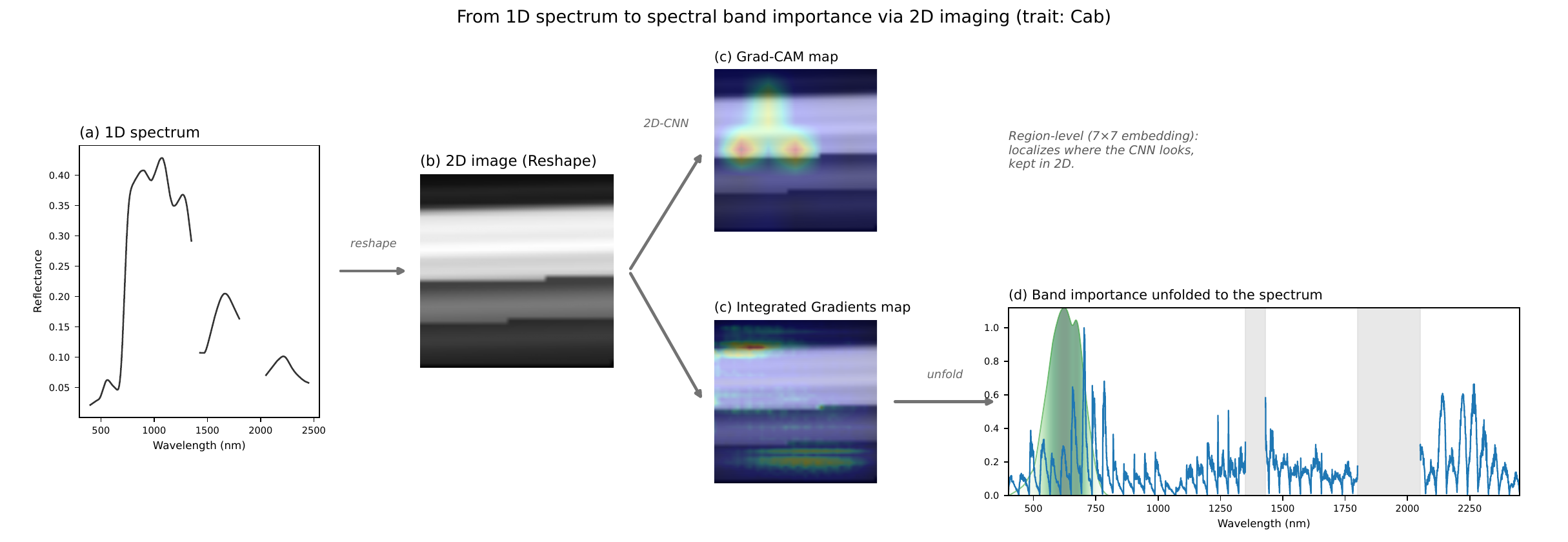}
    \caption{Spectral importance pipeline, illustrated for chlorophyll (C\textsubscript{ab}). A 1D spectrum is folded into a 2D Reshape image and fed to the EfficientNet-B0 model. Grad-CAM and Integrated Gradients then attribute the prediction on that image independently of each other, so neither method takes the other as input. Grad-CAM stays in two dimensions, because it operates on the $7 \times 7$ convolutional embedding and localizes regions rather than individual bands. The Integrated Gradients map is unfolded back to the 1721-band wavelength axis using the band-order-preserving property of the Reshape transform. In panel (d) the green shading marks the PROSAIL theoretical sensitivity and the grey bands mark the removed water-absorption ranges.}
    \label{fig:pipeline}
\end{figure}

We exploited the fact that the Reshape transform preserves band order (pixel $(i,j)$ maps to band $k = 42i + j$) to unfold the IG attribution map back to the 1721-band wavelength axis. The 2D map, therefore, becomes a 1D importance profile directly comparable to classical band selection in remote sensing. We kept Grad-CAM in two dimensions, because it operates on the $7 \times 7$ convolutional embedding and aliases into a comb pattern when forced onto the fine band axis. This pipeline is illustrated in Figure~\ref{fig:pipeline}, where the green shadows show the most important features in the original data.

Finally, we compared the empirical band importance against a physically grounded reference derived from the PROSAIL radiative transfer model using the \texttt{prosail} package \citep{feret2021prospectpro, verhoef2007sail}. For each trait, we swept its corresponding leaf or canopy parameter across the full range in 25 steps while holding the remaining leaf parameters at their \texttt{GreenHyperSpectra} mean. We fixed the structure, soil, and geometry parameters at the values that \citet{cherif2025greenhyperspectra} used to build their look-up table, namely a leaf structure parameter of 1.5, a brown pigment fraction of 0.25, an average leaf angle of 57$^{\circ}$, a hot-spot parameter of 0.01, a soil brightness of 0.8, and a viewing geometry of 0$^{\circ}$ view zenith, 30$^{\circ}$ solar zenith, and 0$^{\circ}$ relative azimuth. We smoothed the simulated spectra, removed the same three water-absorption ranges, and aligned them to the 1721-band axis of the model input. The theoretical sensitivity $S(\lambda)$ of each trait is the coefficient of variation of reflectance across the swept curves at every band, normalized to the zero-to-one range. We used the relative coefficient of variation rather than the raw standard deviation, because the raw deviation peaks on the bright near-infrared plateau for every constituent, an absolute-magnitude artifact that masks the diagnostic absorption features. We then quantified the agreement between the unfolded IG importance and $S(\lambda)$ with the Pearson correlation $r$ and the Spearman correlation $\rho$ over all 1721 bands, and averaged the IG profiles across the three random seeds.


\section{Results}
\label{sec:results}

\subsection{Reflectance Transformation Comparison}
\label{sec:results_transforms}

Table~\ref{tab:transforms} summarizes the performance of all nine 1D-to-2D spectral transformations evaluated with EfficientNet-B0 on the labeled subset of GreenHyperSpectra dataset. Five of them significantly outperformed ($\alpha = 0.05$) the supervised EfficientNet-1D baseline of \citet{cherif2025greenhyperspectra} ($R^2 = 0.587$, the same architecture using the 1D representation), each winning seven to eight of the eight traits: Reshape, Serpentine, and Hilbert, the multi-window spectrogram, and the continuous wavelet transform. The all-band normalized-difference matrix, Gramian Angular Fields, and two-dimensional correlation spectroscopy did not differ significantly from the baseline, and the Markov Transition Field was significantly worse (Figure~S1 of the supplementary material). Among the significant methods, the direct Reshape ranked highest ($R^2 = 0.684 \pm 0.001$, $+0.097$ over the 1D baseline) and showed the lowest variance across random seeds. Serpentine ($R^2 = 0.675 \pm 0.004$) and Hilbert ($R^2 = 0.674 \pm 0.005$) trailed it by less than 0.01, so preserving band adjacency across row boundaries changed almost nothing. The continuous wavelet transform (CWT, $R^2 = 0.641 \pm 0.004$) and multi-window spectrogram ($R^2 = 0.636 \pm 0.025$) followed, and the all-band normalized-difference matrix reached a comparable $R^2 = 0.630 \pm 0.029$ despite encoding every band pair explicitly. Gramian Angular Fields (GAF, $R^2 = 0.609 \pm 0.012$) and two-dimensional correlation spectroscopy (2D-COS, $R^2 = 0.555 \pm 0.077$) formed the lowest tier above the Markov Transition Field, with 2D-COS the most variable across seeds of any transform. Finally, the five-channel composite stacking the best member of each family, namely Reshape, CWT, and NDI, yielded $R^2 = 0.666 \pm 0.017$, below the plain Reshape ($-0.018$) but still significantly above the 1D baseline ($+0.079$, $p = 0.008$, winning all eight traits), so combining complementary encodings did not recover information beyond the single Reshape layout.

\begin{table}[htbp]
\centering
\caption{Comparison of 1D-to-2D spectral transformations for multi-trait retrieval using EfficientNet-B0 on the GreenHyperSpectra dataset. Mean $R^2$ and standard deviation are reported across three random seeds (155, 240, 318). All methods are compared against the supervised EfficientNet-1D baseline of \citet{cherif2025greenhyperspectra} ($R^2 = 0.587$). Bold and underlined fonts show the best- and second-best-fitted models, respectively. CWT=Continuous Wavelet Transform, NDI=all-band Normalized-Difference matrix, GAF=Gramian Angular Fields, 2D-COS=Two-Dimensional Correlation Spectroscopy, MTF=Markov Transition Field. An asterisk ($^{*}$) marks a $\Delta$ significantly different from the 1D baseline (Wilcoxon signed-rank test across the eight traits, $p < 0.05$).}
\label{tab:transforms}
\begin{tabular}{llccr}
\toprule
Transform & Channels & Mean $R^2$ $\pm$ std & $\Delta$ vs.\ 1D \\
\midrule
\textbf{Reshape}     & 1 & \textbf{0.684 $\pm$ 0.001} & \textbf{+0.097$^{*}$} \\
\underline{Serpentine} & 1 & \underline{0.675 $\pm$ 0.004} & \underline{+0.088$^{*}$} \\
Hilbert                & 1 & 0.674 $\pm$ 0.005 & +0.087$^{*}$ \\
CWT                   & 1 & 0.641 $\pm$ 0.004          & +0.054$^{*}$         \\
Spectrogram           & 3 & 0.636 $\pm$ 0.025          & +0.049$^{*}$         \\
NDI                   & 3 & 0.630 $\pm$ 0.029 & +0.043 \\
GAF                   & 2 & 0.609 $\pm$ 0.012          & +0.022         \\
2D-COS                & 2 & 0.555 $\pm$ 0.077          & $-$0.032       \\
MTF                   & 1 & 0.419 $\pm$ 0.022          & $-$0.168$^{*}$        \\
\midrule
Reshape + CWT + NDI & 5 & 0.666 $\pm$ 0.017 & +0.079$^{*}$ \\
\bottomrule
\end{tabular}
\end{table}

\subsection{Per-Trait Analysis}
\label{sec:results_per_trait}

\begin{figure}[!htbp]
    \centering
    \includegraphics[width=\textwidth]{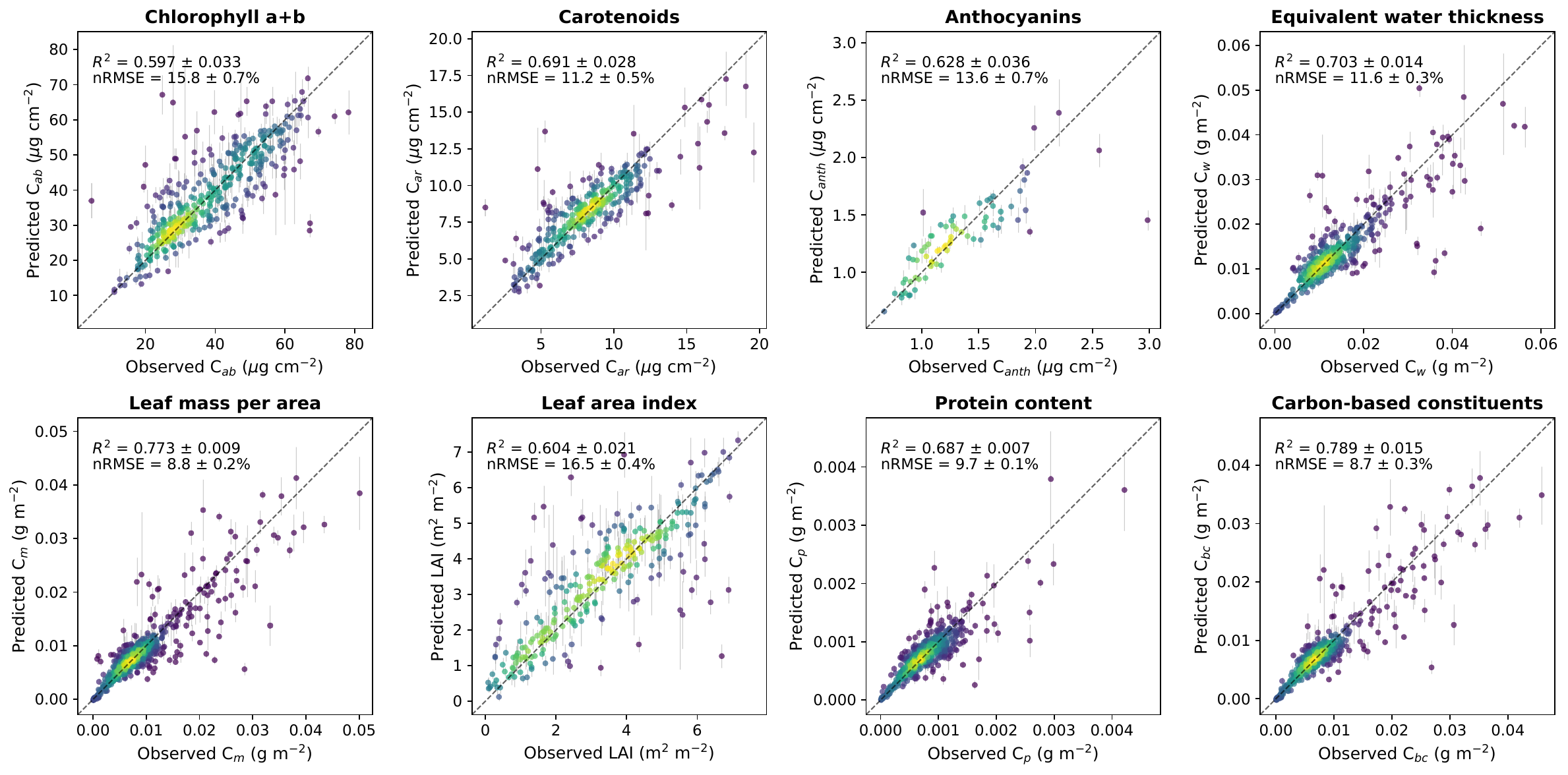}
    \caption{Observed versus predicted trait values for the supervised 2D model (Reshape + EfficientNet-B0) on the GreenHyperSpectra test set ($n = 1{,}127$). Each point is the prediction averaged over the three training seeds, and the grey whisker spans one standard deviation across those seeds. Points are colored by kernel density estimation. The dashed line represents the 1:1 relationship. $R^2$ and nRMSE are given as the mean and standard deviation of the per-seed scores, with nRMSE normalized by the 1st--99th percentile range of the observed values following \citet{cherif2025greenhyperspectra}. Not all test samples have measurements for every trait.}
    \label{fig:scatterplot}
\end{figure}

Table~\ref{tab:per_trait} and Figure~\ref{fig:scatterplot} present the per-trait comparison between the best 2D method (Reshape) and the 1D baseline \citep{cherif2025greenhyperspectra}. The 2D approach improved retrieval accuracy for all eight traits, with per-trait gains ranging from +0.036 for protein content (C\textsubscript{p}) to +0.174 for anthocyanins (C\textsubscript{anth}). The largest gains fell on traits whose retrieval depends on distributed or cross-band information: anthocyanins ($R^2 = 0.628$, $\Delta R^2 = +0.174$), carotenoids ($R^2 = 0.691$, $\Delta R^2 = +0.147$), carbon-based constituents ($R^2 = 0.789$, $\Delta R^2 = +0.110$), and leaf mass per area ($R^2 = 0.773$, $\Delta R^2 = +0.106$). Anthocyanins and carotenoids carry weak, overlapping visible features and are retrieved largely through their covariance with chlorophyll and the red-edge, while carbon-based constituents and leaf mass absorb across several shortwave-infrared windows, so all four benefit from the 2D layout's ability to combine distant bands. Chlorophyll ($R^2 = 0.597$, $\Delta R^2 = +0.080$), equivalent water thickness ($R^2 = 0.703$, $\Delta R^2 = +0.082$), leaf area index ($R^2 = 0.604$, $\Delta R^2 = +0.043$), and protein ($R^2 = 0.687$, $\Delta R^2 = +0.036$) showed smaller but consistent gains, their signal being either a single localized absorption already captured by 1D convolutions or a scattering-dominated canopy response. Chlorophyll content and LAI show greater scatter at higher observed values (Figure~\ref{fig:scatterplot}), reflecting saturation effects well documented in canopy-level reflectance retrievals \citep{cherif2025greenhyperspectra}.

\begin{table}[htbp]
\centering
\caption{Per-trait $R^2$ comparison between the best 2D method (Reshape + EfficientNet-B0) and the supervised 1D baseline \citep{cherif2025greenhyperspectra}. The best value per trait is shown in bold. The asterisk ($^{*}$) on the mean marks a significant overall difference (Wilcoxon signed-rank test across the eight traits, $p = 0.008$). $^{a}$ \citet{cherif2025greenhyperspectra}}
\label{tab:per_trait}
\begin{tabular}{lccr}
\toprule
Trait & Multi-trait 1D$^{a}$ & Reshape 2D & $\Delta$ \\
\midrule
C\textsubscript{ab} (Chlorophyll)    & 0.517 & \textbf{0.597} & +0.080 \\
C\textsubscript{ar} (Carotenoids)    & 0.544 & \textbf{0.691} & +0.147 \\
C\textsubscript{anth} (Anthocyanins) & 0.454 & \textbf{0.628} & +0.174 \\
C\textsubscript{w} (EWT)             & 0.621 & \textbf{0.703} & +0.082 \\
C\textsubscript{m} (LMA)             & 0.667 & \textbf{0.773} & +0.106 \\
LAI                                   & 0.561 & \textbf{0.604} & +0.043 \\
C\textsubscript{p} (Protein)         & 0.651 & \textbf{0.687} & +0.036 \\
C\textsubscript{bc} (CBC)            & 0.679 & \textbf{0.789} & +0.110 \\
\midrule
\textbf{Mean}                         & 0.587 & \textbf{0.684} & \textbf{+0.097$^{*}$} \\
\bottomrule
\end{tabular}
\end{table}

\subsection{Self-Supervised Pretraining}
\label{sec:results_pretraining}

Table~\ref{tab:pretraining} compares the performance of supervised and self-supervised approaches across 1D and 2D representations. 
The 2D supervised approach (Reshape + EfficientNet-B0, $R^2 = 0.684 \pm 0.001$) surpassed all 1D methods, including the best self-supervised MAE-1D FT, by a margin of +0.043. 
The MAE-2D with fine-tuning ($R^2 = 0.667 \pm 0.006$) also outperformed the 1D MAE-FT by +0.026, confirming that the 2D advantage extends to the self-supervised setting. This gain was significant (Wilcoxon $p = 0.039$, winning six of the eight traits). However, the MAE-2D did not surpass the supervised 2D baseline, suggesting that with the current labeled dataset size ($n = 4{,}508$), the pretrained representations do not provide sufficient additional information beyond what direct supervised learning captures.

To assess the quality of the learned representations independently of end-to-end adaptation, we evaluated linear probing: freezing the pretrained MAE-2D encoder and training only a linear regression head (38.6K parameters) on the labeled data. Linear probing achieved $R^2 = 0.646 \pm 0.020$, capturing 97\% of the fine-tuned performance ($R^2 = 0.667$). This result exceeded all 1D baselines reported by \citet{cherif2025greenhyperspectra}, including the MAE-1D fine-tuning result ($R^2 = 0.641$) and the MAE-1D linear probe ($R^2 = 0.466$). The narrow gap between linear probing and fine-tuning (+0.021) indicates that the frozen MAE-2D encoder already encodes most trait-relevant spectral information in its representations, with fine-tuning providing only marginal refinement of the feature space.

\begin{table}[htbp]
\centering
\caption{Comparison of our 2D methods against the corresponding 1D method of \citet{cherif2025greenhyperspectra} for each training regime. All 2D methods use the Reshape transformation. We report the mean $R^2$ across three random seeds for the 2D methods. $\Delta$ is the 2D minus 1D difference for the same regime. Linear probing trains only a regression head on the frozen encoder, while fine-tuning updates the whole encoder. An asterisk ($^{*}$) marks a $\Delta$ significantly different from the 1D counterpart (Wilcoxon signed-rank test across the eight traits, $p < 0.05$). Bold font indicates the best method for each regime. $^{a}$ \citet{cherif2025greenhyperspectra}}
\label{tab:pretraining}
\begin{tabular}{lccr}
\toprule
Training regime & Multi-trait 1D$^{a}$ & Ours 2D & $\Delta$ \\
\midrule
Supervised            & 0.587 & \textbf{0.684 $\pm$ 0.001} & +0.097$^{*}$ \\
MAE fine-tuning       & 0.641 & \textbf{0.667 $\pm$ 0.006} & +0.026$^{*}$ \\
MAE linear probing    & 0.466 & \textbf{0.646 $\pm$ 0.020} & +0.180$^{*}$ \\
\bottomrule
\end{tabular}
\end{table}

\subsection{Out-of-Distribution Generalization}
\label{sec:results_ood}

Table~\ref{tab:ood} shows the cross-dataset evaluation, where every test spectrum comes from a field campaign held out from training. Every transformation lost a large part of its in-distribution accuracy, and the ordering changed. The five-channel composite of Reshape, CWT, and NDI ranked first ($R^2 = 0.333$, $+0.090$ over the supervised 1D baseline), followed by the multi-window spectrogram (0.327) and Serpentine (0.326), while the plain Reshape that led in distribution fell to fourth (0.305). Hilbert dropped the most, from third in distribution to ninth and slightly below the 1D baseline ($-0.010$), while the Markov Transition Field stayed last and significantly below it ($-0.124$, $p = 0.008$). No transformation beat the 1D baseline significantly.
In general, the per-trait analysis of the best four methods showed improvements on five to seven of the eight traits (Table~S5), and the four agreed on six of them. The gains concentrated on the water and dry-matter constituents, equivalent water thickness ($+0.189$ to $+0.257$), carbon-based constituents ($+0.153$ to $+0.180$), leaf mass per area ($+0.139$ to $+0.170$), and protein ($+0.097$ to $+0.154$), which all share a similar shortwave-infrared-driven importance (see section \ref{sec:results_importance}). Carotenoids transferred worse than the 1D baseline for all four ($-0.097$ to $-0.129$) and anthocyanins better ($+0.043$ to $+0.122$), while chlorophyll and leaf area index sat near zero and changed sign from one method to the next.

Fine-tuning the MAE-2D encoder under the same protocol reached an average $R^2$ of 0.153, well below both the supervised 2D model (0.305) and the self-supervised MAE-1D of \citet{cherif2025greenhyperspectra} (0.311). The gap to MAE-1D was significant ($-0.158$, $p = 0.039$) and the 2D encoder won only two of the eight traits (Table~S6). The collapse concentrated on the dry-matter constituents that the supervised model transferred best, leaf mass per area ($-0.352$) and carbon-based constituents ($-0.352$). Under strong domain shift, 1D self-supervision therefore transfers better than our 2D masked autoencoder, so the in-distribution gain of 2D pretraining does not carry to unseen campaigns.

\begin{table}[htbp]
\centering
\caption{Out-of-distribution cross-dataset $R^2$ of every 1D-to-2D transformation under the leave-datasets-out protocol of \citet{cherif2025greenhyperspectra}, averaged over the eight traits and ordered from best to worst. $\Delta$ is the difference from the supervised 1D baseline of \citet{cherif2025greenhyperspectra} ($R^2 = 0.243$), except in the self-supervised block, where it is the difference from their MAE-FR-FT 1D model ($R^2 = 0.311$). Wins counts the traits on which the 2D model exceeds the corresponding 1D reference. Bold and underlined fonts show the best- and second-best-fitted models, respectively. An asterisk ($^{*}$) marks a $\Delta$ significantly different from the 1D reference (Wilcoxon signed-rank test across the eight traits, $p < 0.05$). All values come from a single run. Per-trait nRMSE and $R^2$ are given in Tables~S4 to~S6 of the supplementary material.}
\label{tab:ood}
\begin{tabular}{llcrc}
\toprule
Transform & Channels & $R^2$ & $\Delta$ vs.\ 1D & Wins \\
\midrule
\textbf{Reshape + CWT + NDI} & 5 & \textbf{0.333} & \textbf{+0.090} & 7/8 \\
\underline{Spectrogram}      & 3 & \underline{0.327} & \underline{+0.085} & 5/8 \\
Serpentine            & 1 & 0.326 & +0.084 & 7/8 \\
Reshape               & 1 & 0.305 & +0.062 & 5/8 \\
CWT                   & 1 & 0.286 & +0.043 & 4/8 \\
2D-COS                & 2 & 0.268 & +0.025 & 5/8 \\
NDI                   & 3 & 0.261 & +0.018 & 4/8 \\
GAF                   & 2 & 0.256 & +0.013 & 5/8 \\
Hilbert               & 1 & 0.233 & $-$0.010 & 5/8 \\
MTF                   & 1 & 0.119 & $-$0.124$^{*}$ & 0/8 \\
\midrule
\multicolumn{5}{@{}l}{\textit{Self-supervised regime}}\\
MAE-2D fine-tuning    & 1 & 0.153 & $-$0.158$^{*}$ & 2/8 \\
\bottomrule
\end{tabular}
\end{table}

\subsection{Spectral Importance}
\label{sec:results_importance}

\begin{figure}[!htbp]
    \centering
    \begin{subfigure}{\textwidth}
        \centering
        \begin{overpic}[width=\textwidth]{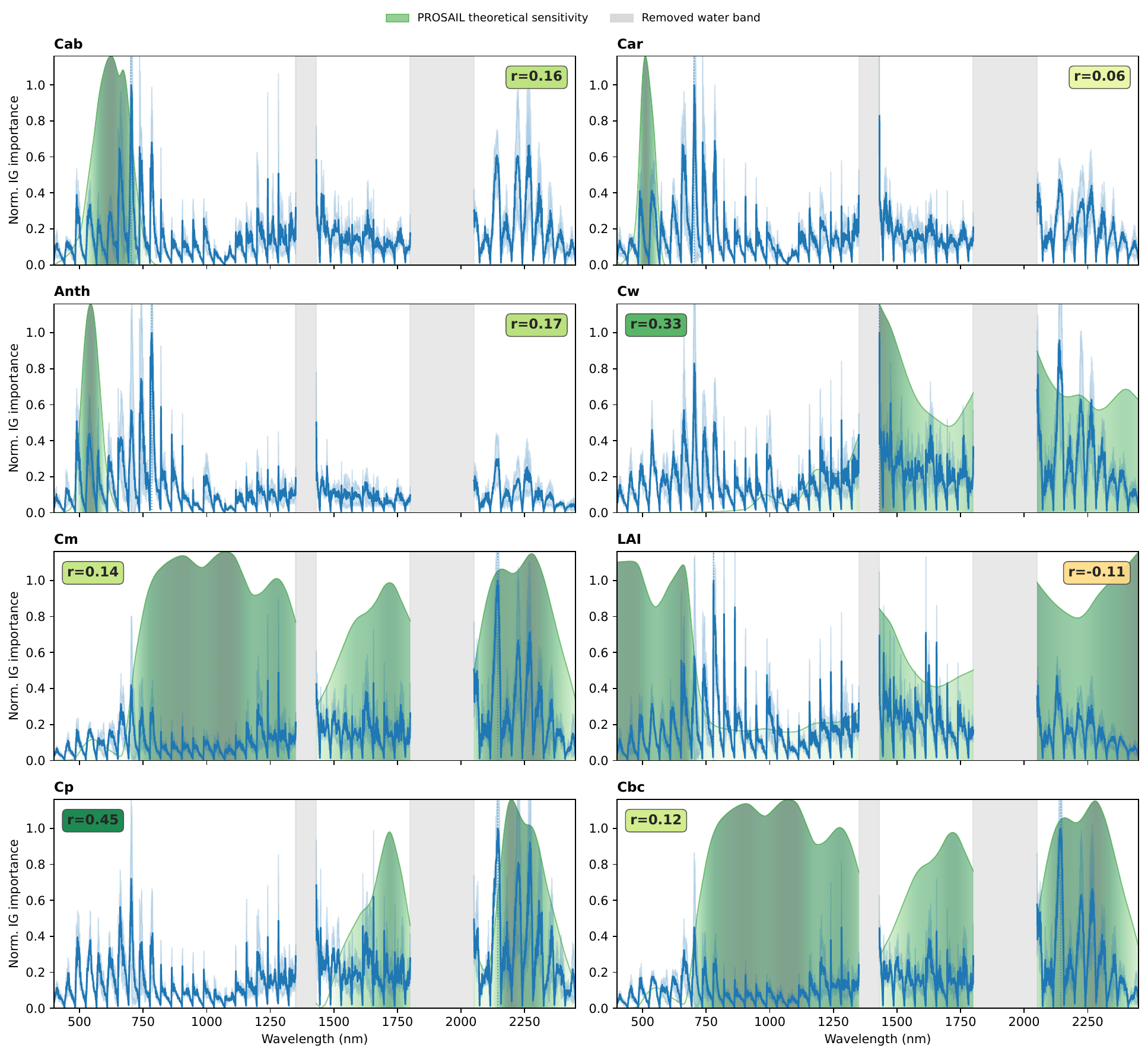}
            \put(0,85){\textbf{(a)}}
        \end{overpic}
        \phantomcaption
        \label{fig:importance_profiles}
    \end{subfigure}
    \par\bigskip
    \begin{subfigure}{\textwidth}
        \centering
        \begin{overpic}[width=\textwidth]{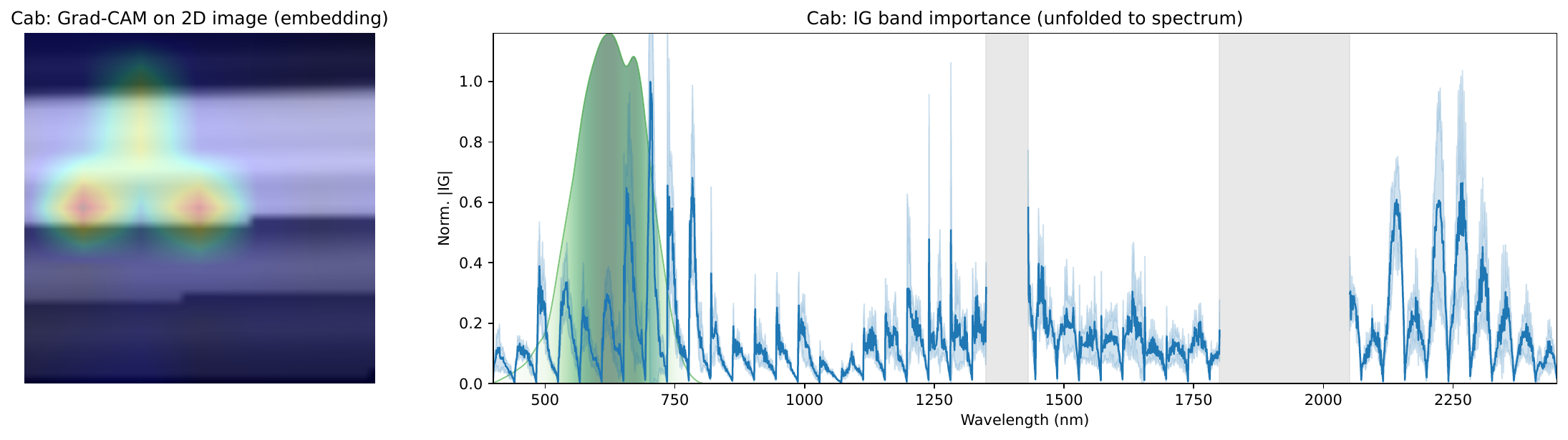}
            \put(1,30){\textbf{(b)}}
        \end{overpic}
        \phantomcaption
        \label{fig:importance_linking}
    \end{subfigure}
    \caption{Spectral importance of the supervised 2D model (Reshape + EfficientNet-B0). (a) Unfolded Integrated Gradients importance profiles for the eight traits, each normalized to its own peak. The blue line is the mean across the three training seeds and the light blue band spans one standard deviation across them. Overlaid are the PROSAIL theoretical spectral sensitivity (green gradient, darker where the constituent is more diagnostic and breaking at the removed water bands), the importance peak, and a per-trait agreement badge. Agreement is the Pearson correlation between the IG importance and the theoretical sensitivity across all 1721 bands. (b) Linking 2D and 1D attribution for chlorophyll (C\textsubscript{ab}). Left, the Grad-CAM relevance over the Reshape image. Right, the Integrated Gradients profile unfolded to the 1721-band wavelength axis. The two panels are independent attributions of the same image, so the right one is not the unfolding of the left.}
    \label{fig:importance}
\end{figure}

Figure~\ref{fig:importance_profiles} shows the unfolded Integrated Gradients profiles, the PROSAIL theoretical sensitivity, and the per-trait agreement between these two. The empirical importance correlated positively with the theoretical sensitivity for seven of the eight traits, but the agreement was modest overall (mean Pearson $r = 0.16$). Protein content showed the strongest agreement ($r = 0.45$), and its empirical peak at 2144\,nm fell close to the theoretical protein N-H overtone at 2198\,nm. Leaf water content followed ($r = 0.33$), with the empirical and theoretical peaks coinciding at the 1431\,nm water edge. Anthocyanins ($r = 0.17$) and chlorophyll a+b ($r = 0.16$) showed weaker positive agreement, and their empirical peaks sat on the red-edge (703 to 784\,nm) rather than at the visible absorption maxima that PROSPECT places at 544 and 624\,nm. Leaf mass per area and carbon-based constituents ($r = 0.14$ and $r = 0.12$) concentrated importance in the shortwave-infrared cellulose and lignin region near 2144\,nm, whereas the theoretical sensitivity peaks in the near-infrared plateau near 1073\,nm. Carotenoids sat near zero ($r = 0.06$), consistent with the model predicting them through their covariance with chlorophyll. Leaf area index was the only trait with negative agreement ($r = -0.11$). Figure~\ref{fig:importance_linking} links an example of the 2D Grad-CAM map and the 1D importance profile for chlorophyll a+b. The per-trait agreement, the Grad-CAM maps, and the linking for the remaining seven traits appear in the supplementary material (Figures~S2 to~S10).

\section{Discussion}
\label{sec:discussion}

\subsection{Why reshaping spectra into images improves trait retrieval}
\label{sec:disc_reshape}

Reshaping the 1D reflectance signal into a 2D image improves multi-trait retrieval by a clear margin in all eight traits, raising the mean $R^2$ from 0.587 for the supervised 1D-CNN to 0.684. Reshape also achieved the highest mean accuracy of the nine transformations we compared, confirming earlier reports that direct reshaping matches or beats more elaborate encodings \citep{hennessy2022reshaping, deev2024spectrum, sharma2019deepinsight} and extending that result to multi-trait regression on heterogeneous field data.

The advantage follows from the multi-scale receptive field that the 2D layout creates. In the $42 \times 42$ image, a single $3 \times 3$ kernel spans about 84\,nm of spectral range through its vertical stride of 42\,nm, whereas an equivalent 1D kernel reaches only about 3\,nm. After a few pooling stages in EfficientNet-B0, the effective receptive field grows to hundreds of nanometers, wide enough to read pigment absorption near 680\,nm, the red-edge near 750\,nm, and shortwave-infrared structural features within one field. A 1D-CNN reaches the same span only with very deep networks or large kernels ($k > 80$), which raises the risk of overfitting on the 4,508 labeled samples \citep{cherif2023spectra}. The reshaped image also carries trait-specific spectral textures, the visually distinct spatial patterns that arise from different biochemical compositions (Figure~\ref{fig:transforms_example}), and 2D convolutions are built to detect exactly this kind of structure.

The per-trait pattern of improvement supports this cross-band mechanism. The largest gains fell on traits whose signal is distributed across the spectrum or carried indirectly by correlated bands. Anthocyanins gained most ($\Delta R^2 = +0.174$) and carotenoids nearly as much ($+0.147$), because the model retrieves both pigments largely through their covariance with chlorophyll and the red-edge rather than through a single visible absorption band \citep{homolova2013review}. Carbon-based constituents and leaf mass per area followed ($+0.110$ and $+0.106$), their dry-matter signal spanning several shortwave-infrared windows \citep{curran1989imaging, feret2021prospectpro}. All four depend on combining bands that lie far apart, exactly what the 2D layout makes available within one receptive field. Chlorophyll and leaf water gained moderately ($+0.080$ and $+0.082$), their features partly reachable by a 1D convolution, while the smallest gains fell on protein ($+0.036$), a single narrow overtone near 2100\,nm, and leaf area index ($+0.043$), a scattering-dominated canopy signal with no sharp absorption. This ordering, from distributed and correlation-driven signals with the largest gains to localized or scattering-driven signals with the smallest, matches the view that the 2D advantage comes from multi-scale cross-band integration.

The ranking of the other transformations reinforces the same point. The continuous wavelet transform (CWT; $R^2 = 0.641$) and the multi-window spectrogram ($R^2 = 0.636$) formed a second tier that keeps some spectral locality through windowed decomposition \citep{addison2017wavelet, shuai2025multichannel}, but placing wavelength on one axis and scale or frequency on the other breaks the wavelength-to-wavelength adjacency that Reshape preserves. These transforms work well for classification tasks that hinge on scale-specific features \citep{ong2025sugarcane, zhuang2023cwt, shuai2025multichannel, zhang2025gaf}, yet continuous regression benefits more from the direct wavelength topology. Gramian Angular Fields (GAF; $R^2 = 0.609$) map reflectance into angular space and group bands by magnitude rather than by wavelength, which discards the physical ordering that ties absorption features to biochemistry \citep{wang2015gaf, ganesan2021gaf}, and both GAF and the Markov Transition Field (MTF) were designed for time-series classification where phase dynamics carry the signal. Two-dimensional correlation spectroscopy ($R^2 = 0.555$, std $= 0.077$) showed the highest variability across seeds, because our single-sample adaptation through windowed segmentation \citep{noda2004twodimensional} perturbs the spectrum in an ad-hoc way that can add noise rather than genuine correlation structure. The MTF ($R^2 = 0.419$) was the only transform below the 1D baseline, since its quantile-based discretization removes the continuous spectral detail that regression depends on \citep{wang2015gaf}. The all-band normalized-difference matrix (NDI; $R^2 = 0.630$) belongs to the same group of abstract encodings, because it replaces the ordered spectrum with an all-pairs index matrix and discards the absolute reflectance level, so even though it encodes every band interaction explicitly, it fell well below the direct layout. Across the whole comparison, the transforms that preserve the natural wavelength ordering and keep continuous values outperform those that remap the spectrum into abstract mathematical spaces \citep{inoue2008ndsi}.


The direct Reshape approach obtained the best results even among layouts that keep the wavelength order intact. The Serpentine and Hilbert-curve variants were designed to remove the row-boundary seam of the raster, where two spectrally adjacent bands land far apart in the image. That seam is not a real bottleneck, because after a few pooling stages the EfficientNet receptive field already spans it, and what the two variants give up is more costly than what they fix. Direct Reshape gives every kernel a regular and globally consistent geometry, a horizontal step of one band and a vertical step of 42 bands, so the same kernel reads the same spectral relationship anywhere in the image. The Serpentine layout reverses every other row, which flips the sign of that vertical step from row to row, while the Hilbert curve follows a fractal path along which no fixed pixel offset maps to a constant spectral distance. Both preserve local adjacency but sacrifice the repeatable stride that the translation-equivariant convolutions exploit, and the Hilbert layout further resamples the spectrum to 4,096 points, which smooths it and removes the exact band positions that the raster preserves. Adjacency preservation therefore buys no accuracy in distribution, while the plain Reshape keeps the lowest variance of any layout. Under the cross-dataset shift, however, that pattern changes. Serpentine held up better than Reshape on held-out campaigns, winning seven of the eight traits against the 1D baseline where Reshape won five, whereas Hilbert fell below that baseline. The alternating raster therefore costs nothing under a domain shift while the fractal path costs a great deal, which points at the resampling to 4,096 points rather than the loss of a repeatable stride as what hurts Hilbert. Together with the poor showing of the all-band normalized-difference matrix, these results show that hand-crafting either band adjacency or band interactions is unnecessary, since the 2D-CNN already recovers the useful inter-band structure from the simplest ordered layout.

These results extend a trend that now reaches across many fields. Researchers have applied it to vibration traces in bearing-fault diagnosis \citep{zhang2025bearing}, electromyography for limb-movement estimation \citep{xia2017emg}, and electroencephalography for emotion recognition \citep{zhao2024eeg}. Spectroscopy has followed the same path. Early deep chemometrics adapted CNNs for vision back down to one dimension and ran them on raw spectra \citep{malek2017cnn}, and one-dimensional models still dominate quantitative work today \citep{ta2023chemometrics, banerjee2024nircnn, cherif2023spectra}. More recent studies lift the spectrum up into two dimensions instead, through spectrograms \citep{shuai2025multichannel}, spider-plot renderings \citep{mokari2025spider}, or Gramian angular fields \citep{zhang2025gaf}, yet these encodings have served classification far more often than regression. Our study departs from this work on two points. We show that the move to two dimensions helps continuous multi-trait regression, the harder and less-explored task, and we find that the plainest image of all, a direct reshape, beats every engineered encoding we tried. That second result runs against the common intuition that a more elaborate transform captures richer structure, and it echoes reports of direct reshaping performing competitively in tabular and spectral deep learning \citep{hennessy2022reshaping, deev2024spectrum, sharma2019deepinsight}.

\subsection{What the model learns from the 2D representation}
\label{sec:disc_importance}

Using Grad-CAM and Integrated Gradients, we found that the model's relevance forms compact 2D clusters. Compactness alone is a generic property of convolutional networks, so on its own it says nothing about the receptive-field mechanism of Section~\ref{sec:disc_reshape}. What matters is where those clusters fall. Once we unfold them to wavelength, the importance of seven of the eight traits agrees with the PROSAIL radiative-transfer sensitivity (Figure~\ref{fig:importance}). The unfolded maps also show how the model reads the image. Horizontal neighbors are adjacent bands, while vertical neighbors lie 42 positions apart, so a single 2D kernel reads both at once and co-weights a visible feature together with its shortwave-infrared overtone about 42 bands away within one receptive field. A 1D kernel of the same size cannot reach that far. Integrated Gradients and Grad-CAM agree across scales, the first at the fine signed band level and the second at the coarse region level, which makes us more confident that the importance reflects learned spectral structure rather than an artifact of a single method.

The traits with a sharp, localized theoretical signature align best. Protein ($r = 0.45$) and leaf water ($r = 0.33$) place their importance almost exactly where radiative transfer predicts, on the shortwave-infrared protein overtone near 2144\,nm and on the 1431\,nm water edge. Leaf water aligns well even though preprocessing removed its strongest bands near 1450 and 1940\,nm, because both the empirical importance and the theoretical sensitivity peak on the surviving 1431\,nm edge, and the model recovers the water signal from that edge. Leaf mass per area, carbon-based constituents, chlorophyll, and anthocyanins also agree positively, with peaks on the red-edge (C\textsubscript{ab} at 703\,nm) and on shortwave-infrared cellulose and lignin features near 2144\,nm, matching constituent absorption reported at the leaf level \citep{Serbin2020-mk, curran1989imaging, feret2021prospectpro}. 
Carotenoids ($r = 0.06$) and leaf area index ($r = -0.11$) show the weakest agreement, yet the model predicts both accurately, so the gap reflects where the model looks rather than how well it performs. The multi-trait model recovers carotenoids through their strong covariance with chlorophyll, leaning on the chlorophyll red-edge at 703\,nm instead of the weak carotenoid feature at 440--520\,nm, a shortcut that already emerged in the joint retrieval of \citet{cherif2023spectra}. Leaf area index is a structural canopy property whose reflectance signal arises from scattering and the red-edge rather than from a single narrow absorption, so the model concentrates on the red-edge near 779\,nm even where the theoretical sensitivity peaks elsewhere. Canopy structure and soil background further broaden and shift leaf-level absorption features, so the most predictive bands need not coincide with the leaf-level maxima \citep{homolova2013review}. For leaf mass per area and carbon-based constituents, the theoretical sensitivity peaks in the near-infrared plateau near 1073\,nm, where canopy multiple scattering dominates, yet the model concentrates on the shortwave-infrared cellulose and lignin absorption near 2144\,nm, indicating that it relies on diagnostic constituent absorption rather than the higher-variance scattering response. Anthocyanins follow the same red-edge pattern as carotenoids. Both PROSPECT-PRO and leaf-level optics place the anthocyanin absorption near 550\,nm \citep{feret2021prospectpro, gitelson2001anthocyanin}, yet the model leans on the red-edge, which also carries a documented anthocyanin signal exploited by reflectance indices \citep{gitelson2009anthocyanin}, rather than on the green absorption itself.

Our interpretation approach also connects the model to a growing use of explainable AI in remote sensing. Grad-CAM and related attribution methods now routinely open the black box of vision networks in fields as varied as wildfire-spread modeling \citep{zhou2025wildfire} and crop phenotyping \citep{danilevicz2025xai}, although applying them to geospatial data raises its own difficulties around scale and reference choice \citep{xing2023geoai}. Most of this work explains classification maps in the spatial domain. We instead fold the attribution back onto the spectral axis, which turns a 2D relevance map into a 1D band-importance profile and lets us compare it directly against radiative-transfer theory. Other methods could read a spectral model in different ways. Model-agnostic attributions such as SHAP and occlusion work on any predictor, and spectral-specific schemes include variable importance in partial least squares, explainable-AI-guided band selection in hyperspectral imaging \citep{khatun2026xai}, attention weights learned inside the network \citep{kong2025maize}, and 2D correlation analysis of the learned features \citep{contreras2025xai2dcos}. We chose Integrated Gradients and Grad-CAM because they need no change to the architecture, agree with each other across scales, and map cleanly back to wavelength.


The cross-dataset test sharpens this reading of what the model uses. The 2D reshape transferred best on the water and dry-matter traits, exactly the constituents whose importance sat on diagnostic shortwave-infrared absorption. The best four models transferred worst on carotenoids and leaf area index, whose importance leaned on the red-edge shoulder and on broad scattering rather than on a constituent absorption band. A red-edge shortcut is efficient within one dataset but tied to the local mix of species, canopy structure, and background, so it generalizes less well to unseen sensors and ecosystems, whereas the shortwave-infrared absorption of cellulose, lignin, and water reflects constituent chemistry that stays stable across campaigns. Anthocyanins are the exception, gaining despite the same indirect retrieval, although that trait carries the widest spread across subsamples of any trait, so we do not read much into its margin. The traits the model reads through genuine absorption are therefore also the traits whose accuracy survives the domain shift, which ties the interpretation analysis directly to the generalization result.

\subsection{Self-supervised pretraining in the 2D framework}
\label{sec:disc_pretraining}

Self-supervised pretraining helped less than the change in representation. Moving from the 1D to the 2D image raised the supervised $R^2$ by 0.097, whereas pretraining a 1D masked autoencoder on 139{,}000 unlabeled spectra raised the 1D model by only 0.054 \citep{cherif2025greenhyperspectra}. The gain therefore came from the representation, not the learning strategy.

The 2D representation also outperformed its 1D counterpart under self-supervised pretraining. MAE-2D fine-tuning ($R^2 = 0.667 \pm 0.006$) beat MAE-1D fine-tuning by 0.026, so the 2D advantage held in the pretrained setting as well. It did not, however, beat the supervised 2D model ($-0.017$), unlike natural-image domains where masked-autoencoder pretraining usually helps \citep{he2022mae}. Under the cross-dataset shift, MAE-2D fine-tuning dropped to 0.153, below both our supervised 2D model and the 1D masked autoencoder (0.311, $p = 0.039$). Masked reconstruction on 139K spectra therefore builds features that suit the pretraining distribution and transfer worse than 1D self-supervision once the campaign changes. The frozen encoder was nonetheless strong. Linear probing, which trains only a 38.6K-parameter head, reached $R^2 = 0.646 \pm 0.020$ and beat every 1D baseline including MAE-1D fine-tuning. It recovered 97\% of the fine-tuned performance, which shows that the frozen encoder already holds most of the trait-relevant spectral information.

This pattern matches a broader trend in self-supervised learning. Masked-autoencoder and contrastive pretraining deliver their largest gains when labeled data are scarce, a result reported across plant phenotyping \citep{nagasubramanian2022}, medical imaging \citep{jiang2024ssl}, and image classification with small training sets \citep{wittscher2022ssl}. Our moderate labeled set of 4,508 samples, together with a 2D representation that already separates the traits well, leaves little room for pretraining to add information, which is why MAE-2D did not beat the supervised 2D model.

A structural asymmetry explains why pretraining pays off so differently across the two representations. Supervision on the 2D image already exposes long-range band pairs, because a single kernel spans a vertical step of 42 bands (Section~\ref{sec:disc_reshape}). The masked autoencoder reaches those pairs only partly: at $16 \times 16$ pixels, one patch covers about three bands on each of three rows, so it does mix runs separated by 42 bands, but it learns that mixing for reconstruction rather than for trait prediction. A 1D masked autoencoder has no such access at all, since its patches are contiguous runs of the raw spectrum. That is why 1D pretraining still adds a genuine 0.054, while 2D pretraining adds nothing ($-0.017$).

Three further factors likely account for the supervised 2D model staying ahead of MAE-2D fine-tuning. First, EfficientNet-B0 brings spatial inductive biases such as local connectivity and translation equivariance that outperform the learned attention of a vision transformer when only 4,508 labeled samples are available \citep{dosovitskiy2021vit}. Second, reshaped spectral images are smooth and continuous rather than complex like natural photographs, so the reconstruction task converges within about ten epochs (val\_loss near 0.0001) and gives the encoder little pressure to learn deep trait-discriminative features. Third, the 4,508 samples span diverse ecosystems and may already cover most of the learnable spectral-to-trait relationships. In practice a reshape combined with a standard CNN reaches state-of-the-art accuracy without the cost of pretraining on a large unlabeled corpus.

\subsection{Implications for trait-based ecology and remote sensing}
\label{sec:disc_implications}

The gains translate into concrete benefits for trait-based ecology. Carbon-based constituents and leaf mass per area, which govern a plant's dry-matter investment, both improved substantially ($\Delta R^2 = +0.110$ and $+0.106$), which matters directly for carbon stocks, litter decomposition, and nutrient cycling \citep{kattge2020try}. Better retrieval of these traits strengthens non-destructive estimation of the dry-matter economics that anchor the leaf economics spectrum \citep{wright2004les}, and supports more reliable mapping of ecosystem carbon pools and decomposition dynamics for Earth system models \citep{homolova2013review}.

Predicting all traits with one model preserves the covariance structure among them, which functional diversity metrics such as FDis, FRic, FEve, and Rao's Q depend on \citep{villeger2008fd, laliberte2010fdis}. Fitting a separate model per trait can distort those correlations and bias the resulting diversity estimates. Because the 2D approach improves all eight traits at once, the functional trait spaces built from its predictions are both more accurate and more internally consistent than those from 1D baselines, which benefits biodiversity monitoring programs that rely on spectral trait mapping \citep{schweiger2018plant, binh2025mangrove, weiss2020remote}.

The method is also easy to deploy. Reshape needs no domain-specific tuning, so there are no wavelet scales, spectrogram windows, quantile bins, or perturbation parameters to set, and it takes roughly six lines of code. Its stability across the three random seeds (std $= 0.001$) gives reproducible results without hyperparameter sensitivity. This contrasts with the run-to-run instability that pretraining-based methods still report on the same data \citep{cherif2025greenhyperspectra}, and it lowers the barrier for ecologists and remote sensing practitioners who need strong predictions without signal-processing expertise. This simplicity suits the data streams from recent and upcoming hyperspectral missions such as EnMAP \citep{guanter2015enmap}, PRISMA \citep{verrelst2021prisma}, and SBG and EMIT \citep{cawse2021sbg, green2020emit}, which are producing continental to global archives of contiguous spectra across 400 to 2500\,nm. A method that stays simple and stable across data from 50 heterogeneous sources is well placed for operational use on these archives. Beyond raw sensor archives, this simplicity fits the operational pipelines that now underpin biodiversity policy. The IPBES framework calls for monitoring functional-trait diversity across scales \citep{diaz2015ipbes}, and plant traits sit directly within the species-traits class of the Essential Biodiversity Variables coordinated by GEO BON \citep{pereira2013ebv, jetz2016plantdiversity}. Platforms such as BON-in-a-Box now assemble these variables into policy indicators through open and reproducible pipelines \citep{Griffith2026-vs}, which reward methods that run without per-dataset tuning and return stable predictions. A reshape paired with a standard CNN meets that need, so it can slot into an EBV pipeline as a trait-retrieval step with little adaptation.

The self-supervised results open a further path for undersampled biomes. The MAE-2D linear probe reaches $R^2 = 0.646$ with only 38.6K trainable parameters, which suits trait prediction in data-scarce systems such as tropical forests, tundra, and peatlands that remain underrepresented in trait databases \citep{kattge2020try, zhang2025ppadanet}, where self-supervised pretraining delivers its largest gains \citep{cherif2025greenhyperspectra}. A frozen pretrained encoder paired with a lightweight linear head could map traits with very little local calibration data, in line with the broader move toward foundation models for remote sensing \citep{binfaruk2025terramae}.

\subsection{Limitations and future directions}
\label{sec:disc_limitations}

Several limitations frame these results. We compared all transformations on a single backbone, EfficientNet-B0, so the rankings could shift with other architectures, although \citet{cherif2023spectra} found this network well matched to the present data scale. We also fixed the image size at $224 \times 224$, and since the reshape topology depends on resolution, other sizes may favor different transforms, which warrants a systematic study. Our single-sample version of two-dimensional correlation spectroscopy approximates the multi-sample formulation of \citet{noda2004twodimensional} through windowed segmentation, and its weak performance may partly reflect that approximation. On domain shift with a cross-dataset evaluation, every model lost a large fraction of its accuracy on held-out campaigns, which leaves substantial room for methods that close the domain gap. That evaluation also rests on a single training run per representation, and it is sensitive to choices outside the protocol. Replacing the spectral augmentation of \citet{cherif2025greenhyperspectra} with the image-space augmentation we use elsewhere moved the supervised average by 0.032 and the pretrained average by 0.120, both comparable to the 1D-to-2D effect itself, so the out-of-distribution margins should be read as indicative rather than as a stable ranking. We also evaluated only full-range spectra, whereas many satellite and drone sensors record the visible and near-infrared alone. The reshape geometry ties directly to the band count, so a half-range input of about 500 bands folds into a smaller and differently connected image and loses the visible-to-shortwave-infrared co-weighting the model relies on, which leaves half-range performance \citep{cherif2025greenhyperspectra} an open question. Finally, each image encodes one rearranged spectrum with no spatial neighbors, so the point-level design carries none of the spatial autocorrelation that can weaken image-based spectral datasets \citep{cherif2025greenhyperspectra}, though it also gives the model no spatial context, whereas airborne and satellite imagery add mixed pixels and atmospheric effects that we did not address here. Reshape was also chosen as the representative 2D method on the same fixed test set used to report its accuracy, so its headline $R^2$ may be marginally optimistic, although the small and consistent margin over the next layouts makes a large selection effect unlikely.

The reshape geometry is arbitrary, which limits how finely we can read the importance maps. The model zero-pads the spectrum and resizes it bilinearly from $42 \times 42$ to $224 \times 224$, which smears each band across about five pixels. The vertical 42-band adjacency comes from the side-42 fold rather than from physics, so part of the cross-row structure reflects the layout, and the row-wrap boundaries place spectrally adjacent bands far apart in the image. The unfolded importance therefore has limited spectral resolution, and the apparent peaks can shift by a few bands, so we read the importance maps as approximate.

The layout also leaves a periodic imprint on the unfolded profiles, a ripple whose period matches the 42-band fold and which carries between 19 and 41 percent of the profile variance depending on the trait. The ripple arises because the last band of every row falls on the right edge of the $224 \times 224$ image, where we measured the input gradient dropping to 6 percent of its interior value. Two effects combine there. The convolutions pad the image border, so the receptive field of the outermost pixels is half empty, and the network additionally learns to ignore that corner because the 43 zero-padding cells of the reshape sit in it and never carry information. The model therefore underweights roughly one band in every 42, together with its immediate neighbors, which is a genuine property of the representation rather than an artifact of the attribution. It does not affect the results we report, because the importance peak of all eight traits falls in an interior column of the grid, between columns 1 and 28 of 42, well away from the attenuated edge. This also explains why the Serpentine and Hilbert layouts did not improve accuracy despite preserving band adjacency, since both still place bands on the image border. Padding the reshape by reflection instead of with zeros, or replicating the border during convolution, would remove the dead corner and is worth testing.

These limitations point to clear next steps. The effect of image size and reshape topology deserves direct study, since different row widths create different cross-band neighborhoods that may favor different traits. Harder self-supervised objectives are also worth exploring, including higher mask ratios, masking of whole spectral regions, and richer inputs such as wavelet scalograms in place of the reshape. Pretraining larger 2D encoders on massive spectral archives could yield general-purpose feature extractors in the spirit of foundation models. Combining reshaped spectral images with spatial context from imaging spectrometers or with canopy structure from LiDAR offers a route to multi-modal fusion.

\section{Conclusion}

We showed that turning one-dimensional hyperspectral reflectance into a two-dimensional image improves multi-trait retrieval with convolutional neural networks. A direct reshape of the spectrum, the simplest of the nine transformations we tested, gave the best accuracy ($R^2 = 0.684$) and improved all eight traits over the one-dimensional baseline ($+0.097$) and over every more elaborate encoding. The advantage came from the representation itself rather than from architectural complexity or pretrained ImageNet weights, because the two-dimensional layout lets a single kernel integrate spectrally distant bands within one receptive field. Hand-crafting either band adjacency or band interactions did not help, since the network already recovers the useful inter-band structure from the plain ordered layout. Attribution with Integrated Gradients and Grad-CAM showed that for traits with sharp absorption features, such as protein and leaf water, the model reads genuine constituent absorption in line with radiative-transfer sensitivity, while for broad or structural traits it leans on correlated red-edge and near-infrared features.

Under self-supervised pretraining, the two-dimensional representation again outperformed its one-dimensional counterpart, and a frozen encoder with a lightweight linear head already beat every one-dimensional baseline, which points to its value where labeled data are scarce. Because a reshape paired with a standard network is simple, stable, and reproducible, it suits the operational pipelines that now can convert spectra into trait distributions for global monitoring. Testing the framework across sensor ranges and out-of-distribution domains, and scaling two-dimensional encoders toward foundation models, are the natural next steps.

\section*{CRediT authorship contribution statement}

\textbf{Javier Lopatin:} Conceptualization, Methodology, Software, Formal analysis, Investigation, Data curation, Visualization, Writing -- original draft, Writing -- review \& editing. \textbf{Teja Kattenborn, Eya Cherif and Sebastián Moreno:} Writing -- review \& editing.

\section*{Declaration of competing interest}

The authors declare that they have no known competing financial interests or personal relationships that could have appeared to influence the work reported in this paper.

\section*{Data availability}

The GreenHyperSpectra dataset used in this study is publicly available at \url{https://huggingface.co/datasets/Avatarr05/GreenHyperSpectra} \citep{cherif2025greenhyperspectra}. All code to reproduce the transformations, models, and analyses is available at \url{https://github.com/JavierLopatin/Trait_2DCNN}.

\section*{Funding}

This work was supported by the Agencia Nacional de Investigación y Desarrollo (ANID), Chile [grant number 11241088]. The funder had no role in the study design, in the collection, analysis and interpretation of data, in the writing of the report, or in the decision to submit the article for publication.

\section*{Declaration of generative AI and AI-assisted technologies in the manuscript preparation process}

During the preparation of this work the authors used Claude (Anthropic) to support the development of the analysis code. After using this tool, the authors reviewed and edited the content as needed and take full responsibility for the content of the published article.

\bibliographystyle{elsarticle-harv}
\bibliography{References}

\clearpage
\setcounter{table}{0}
\setcounter{figure}{0}
\renewcommand{\thetable}{S\arabic{table}}
\renewcommand{\thefigure}{S\arabic{figure}}
\input{supplementary_body}

\end{document}

%% file: supplementary_body.tex
\section*{Supplementary Data}

\begin{table}[htbp]
\centering
\caption{Mean nRMSE (\%) by 1D-to-2D transformation on the in-distribution GreenHyperSpectra test set (mean $\pm$ standard deviation over three seeds), complementing the $R^2$ ranking of the main-text transform comparison. nRMSE is normalized by the 1st--99th percentile range of the observed values, following Cherif et al. (2025), so that lower is better. $\Delta$ is the difference from their supervised 1D baseline (nRMSE 13.70), where a negative value means a lower error than the 1D baseline. The best value is in bold and the composite is listed below the rule.}
\label{tab:transforms_nrmse}
\begin{tabular}{lcr}
\toprule
Transform & Mean nRMSE (\%) & $\Delta$ vs.\ 1D \\
\midrule
Reshape          & \textbf{12.00 $\pm$ 0.02} & $-$1.70 \\
Serpentine       & 12.15 $\pm$ 0.09 & $-$1.55 \\
Hilbert          & 12.16 $\pm$ 0.09 & $-$1.54 \\
CWT              & 12.76 $\pm$ 0.06 & $-$0.94 \\
Spectrogram      & 12.82 $\pm$ 0.43 & $-$0.88 \\
NDI              & 12.96 $\pm$ 0.50 & $-$0.74 \\
GAF              & 13.30 $\pm$ 0.14 & $-$0.40 \\
2D-COS           & 14.12 $\pm$ 0.97 & +0.42 \\
MTF              & 16.18 $\pm$ 0.25 & +2.48 \\
\midrule
Reshape + CWT + NDI  & 12.34 $\pm$ 0.30 & $-$1.36 \\
Cherif 1D (baseline) & 13.70 & -- \\
\bottomrule
\end{tabular}
\end{table}

\begin{table}[htbp]
\centering
\caption{Per-trait in-distribution nRMSE (\%) of the best model (Reshape plus EfficientNet-B0, mean over three seeds) against the supervised 1D baseline of Cherif et al. (2025), complementing the per-trait $R^2$ comparison of the main text. Lower is better and the lower value per trait is in bold. $\Delta$ is the 2D minus 1D difference.}
\label{tab:reshape_nrmse}
\begin{tabular}{lccr}
\toprule
Trait & Cherif 1D & Reshape 2D & $\Delta$ \\
\midrule
C\textsubscript{ab} (Chlorophyll)    & 17.34 & \textbf{15.83} & $-$1.51 \\
C\textsubscript{ar} (Carotenoids)    & 13.65 & \textbf{11.23} & $-$2.42 \\
C\textsubscript{anth} (Anthocyanins) & 16.46 & \textbf{13.58} & $-$2.89 \\
C\textsubscript{w} (EWT)             & 13.10 & \textbf{11.59} & $-$1.52 \\
C\textsubscript{m} (LMA)             & 10.67 & \textbf{8.83}  & $-$1.84 \\
LAI                                   & 17.49 & \textbf{16.50} & $-$0.98 \\
C\textsubscript{p} (Protein)         & 10.23 & \textbf{9.74}  & $-$0.48 \\
C\textsubscript{bc} (CBC)            & 10.63 & \textbf{8.66}  & $-$1.98 \\
\midrule
\textbf{Mean}                         & 13.70 & \textbf{12.00} & $-$1.70 \\
\bottomrule
\end{tabular}
\end{table}

\begin{table}[htbp]
\centering
\caption{In-distribution nRMSE (\%) by training regime, complementing the $R^2$ comparison of the main-text pretraining table. All 2D methods use the Reshape transformation (mean $\pm$ standard deviation over three seeds). The 1D values are the corresponding nRMSE of Cherif et al. (2025) for the same regime. Lower is better and the lower value per regime is in bold. $\Delta$ is the 2D minus 1D difference.}
\label{tab:pretraining_nrmse}
\begin{tabular}{lccr}
\toprule
Training regime & Cherif 1D & Ours 2D & $\Delta$ \\
\midrule
Supervised            & 13.70 & \textbf{12.00 $\pm$ 0.02} & $-$1.70 \\
MAE fine-tuning       & 12.78 & \textbf{12.30 $\pm$ 0.13} & $-$0.48 \\
MAE linear probing    & 15.50 & \textbf{12.62 $\pm$ 0.27} & $-$2.87 \\
\bottomrule
\end{tabular}
\end{table}

\begin{table}[htbp]
\centering
\caption{Out-of-distribution cross-dataset nRMSE (\%) under the leave-datasets-out protocol (Cherif et al., 2025), split by training regime, complementing the $R^2$ values reported for the same experiment in the main text. The 1D values are as reported by Cherif et al. (2025); the 2D values use the Reshape transformation from a single run. $\Delta$ is the 2D minus 1D difference, and the lower (better) value per trait is shown in bold.}
\label{tab:ood_nrmse}
\begin{tabular}{lccr}
\toprule
Trait & Cherif 1D & Reshape 2D & $\Delta$ \\
\midrule
\multicolumn{4}{@{}l}{\textit{Supervised}}\\
C\textsubscript{ab} (Chlorophyll)    & \textbf{19.17} & 20.87 & +1.70 \\
C\textsubscript{ar} (Carotenoids)    & \textbf{19.18} & 20.56 & +1.38 \\
C\textsubscript{anth} (Anthocyanins) & 23.16 & \textbf{22.02} & $-$1.14 \\
C\textsubscript{w} (EWT)             & 25.22 & \textbf{17.76} & $-$7.46 \\
C\textsubscript{m} (LMA)             & 14.24 & \textbf{12.95} & $-$1.29 \\
LAI                                   & \textbf{22.98} & 23.71 & +0.73 \\
C\textsubscript{p} (Protein)         & \textbf{17.07} & 17.44 & +0.37 \\
C\textsubscript{bc} (CBC)            & 14.82 & \textbf{13.08} & $-$1.74 \\
\cmidrule(lr){1-4}
\textbf{Mean}                         & 19.48 & \textbf{18.55} & $-$0.93 \\
\midrule
\multicolumn{4}{@{}l}{\textit{MAE fine-tuning}}\\
C\textsubscript{ab} (Chlorophyll)    & \textbf{20.50} & 20.84 & +0.34 \\
C\textsubscript{ar} (Carotenoids)    & \textbf{19.37} & 20.83 & +1.46 \\
C\textsubscript{anth} (Anthocyanins) & 22.42 & \textbf{21.68} & $-$0.74 \\
C\textsubscript{w} (EWT)             & 24.02 & \textbf{17.93} & $-$6.08 \\
C\textsubscript{m} (LMA)             & \textbf{12.47} & 13.40 & +0.94 \\
LAI                                   & \textbf{20.84} & 22.05 & +1.21 \\
C\textsubscript{p} (Protein)         & \textbf{16.07} & 17.97 & +1.91 \\
C\textsubscript{bc} (CBC)            & \textbf{12.91} & 13.70 & +0.79 \\
\cmidrule(lr){1-4}
\textbf{Mean}                         & 18.57 & \textbf{18.55} & $-$0.02 \\
\bottomrule
\end{tabular}
\end{table}

\begin{table}[htbp]
\centering
\caption{Out-of-distribution cross-dataset $R^2$ per trait for the four best-performing 1D-to-2D transformations under the leave-datasets-out protocol (Cherif et al., 2025), complementing the trait-averaged values reported in the main text. The 1D column holds the supervised EfficientNet-1D values reported by Cherif et al. (2025); the 2D values come from a single run each. The highest (better) value per trait is shown in bold. Wins counts the traits on which each 2D model exceeds the 1D baseline.}
\label{tab:ood_r2_pertrait}
\begin{tabular}{lccccc}
\toprule
Trait & Cherif 1D & Reshape+CWT+NDI & Spectrogram & Serpentine & Reshape \\
\midrule
C\textsubscript{ab} (Chlorophyll)     & 0.362 & \textbf{0.379} & 0.336 & 0.378 & 0.340 \\
C\textsubscript{ar} (Carotenoids)     & \textbf{0.181} & 0.065 & 0.070 & 0.084 & 0.052 \\
C\textsubscript{anth} (Anthocyanins)  & 0.055 & 0.098 & 0.161 & 0.143 & \textbf{0.177} \\
C\textsubscript{w} (EWT)              & 0.193 & 0.413 & \textbf{0.450} & 0.420 & 0.382 \\
C\textsubscript{m} (LMA)              & 0.446 & 0.610 & \textbf{0.616} & 0.607 & 0.585 \\
LAI                                   & 0.074 & \textbf{0.134} & 0.073 & 0.076 & $-$0.005 \\
C\textsubscript{p} (Protein)          & 0.183 & \textbf{0.337} & 0.290 & 0.280 & 0.307 \\
C\textsubscript{bc} (CBC)             & 0.449 & \textbf{0.629} & 0.625 & 0.623 & 0.602 \\
\cmidrule(lr){1-6}
\textbf{Mean}                         & 0.243 & \textbf{0.333} & 0.327 & 0.326 & 0.305 \\
\textbf{Wins vs.\ 1D}                 & -- & 7/8 & 5/8 & 7/8 & 5/8 \\
\bottomrule
\end{tabular}
\end{table}

\begin{table}[htbp]
\centering
\caption{Out-of-distribution cross-dataset $R^2$ per trait under self-supervised pretraining, comparing the MAE-FR fine-tuned 1D model of Cherif et al. (2025) with our MAE-2D encoder fine-tuned on the Reshape representation. Values come from a single run. $\Delta$ is the 2D minus 1D difference, and the higher (better) value per trait is shown in bold.}
\label{tab:ood_r2_mae}
\begin{tabular}{lccr}
\toprule
Trait & Cherif MAE-1D & MAE-2D & $\Delta$ \\
\midrule
C\textsubscript{ab} (Chlorophyll)     & 0.271 & \textbf{0.284} & +0.013 \\
C\textsubscript{ar} (Carotenoids)     & 0.165 & \textbf{0.214} & +0.049 \\
C\textsubscript{anth} (Anthocyanins)  & \textbf{0.112} & $-$0.031 & $-$0.143 \\
C\textsubscript{w} (EWT)              & \textbf{0.280} & 0.192 & $-$0.088 \\
C\textsubscript{m} (LMA)              & \textbf{0.575} & 0.223 & $-$0.352 \\
LAI                                   & \textbf{0.229} & 0.046 & $-$0.183 \\
C\textsubscript{p} (Protein)          & \textbf{0.275} & 0.068 & $-$0.207 \\
C\textsubscript{bc} (CBC)             & \textbf{0.582} & 0.230 & $-$0.352 \\
\cmidrule(lr){1-4}
\textbf{Mean}                         & \textbf{0.311} & 0.153 & $-$0.158 \\
\bottomrule
\end{tabular}
\end{table}

\begin{table}[htbp]
\centering
\caption{Effect of the min-max scaling scope on in-distribution accuracy. Per-image scaling normalizes each image by its own minimum and maximum, while global scaling applies a single minimum and maximum fitted on the training set. Values are the mean $R^2$ and the standard deviation across three random seeds on the GreenHyperSpectra test set, using the Reshape transform with EfficientNet-B0. $\Delta$ is global minus per-image, and the higher (better) value per trait is shown in bold.}
\label{tab:scaling_scope}
\begin{tabular}{lccr}
\toprule
Trait & Per-image & Global & $\Delta$ \\
\midrule
C\textsubscript{ab} (Chlorophyll)     & \textbf{0.597} $\pm$ 0.033 & 0.566 $\pm$ 0.032 & $-$0.031 \\
C\textsubscript{ar} (Carotenoids)     & \textbf{0.691} $\pm$ 0.028 & 0.619 $\pm$ 0.041 & $-$0.072 \\
C\textsubscript{anth} (Anthocyanins)  & \textbf{0.628} $\pm$ 0.036 & 0.585 $\pm$ 0.084 & $-$0.044 \\
C\textsubscript{w} (EWT)              & \textbf{0.703} $\pm$ 0.014 & 0.660 $\pm$ 0.037 & $-$0.043 \\
C\textsubscript{m} (LMA)              & \textbf{0.773} $\pm$ 0.009 & 0.742 $\pm$ 0.028 & $-$0.031 \\
LAI                                   & \textbf{0.604} $\pm$ 0.021 & 0.565 $\pm$ 0.042 & $-$0.039 \\
C\textsubscript{p} (Protein)          & \textbf{0.687} $\pm$ 0.007 & 0.664 $\pm$ 0.031 & $-$0.023 \\
C\textsubscript{bc} (CBC)             & \textbf{0.789} $\pm$ 0.015 & 0.763 $\pm$ 0.020 & $-$0.026 \\
\cmidrule(lr){1-4}
\textbf{Mean}                         & \textbf{0.684} & 0.646 & $-$0.039 \\
\bottomrule
\end{tabular}
\end{table}

\begin{figure}[htbp]
    \centering
    \includegraphics[width=0.9\textwidth]{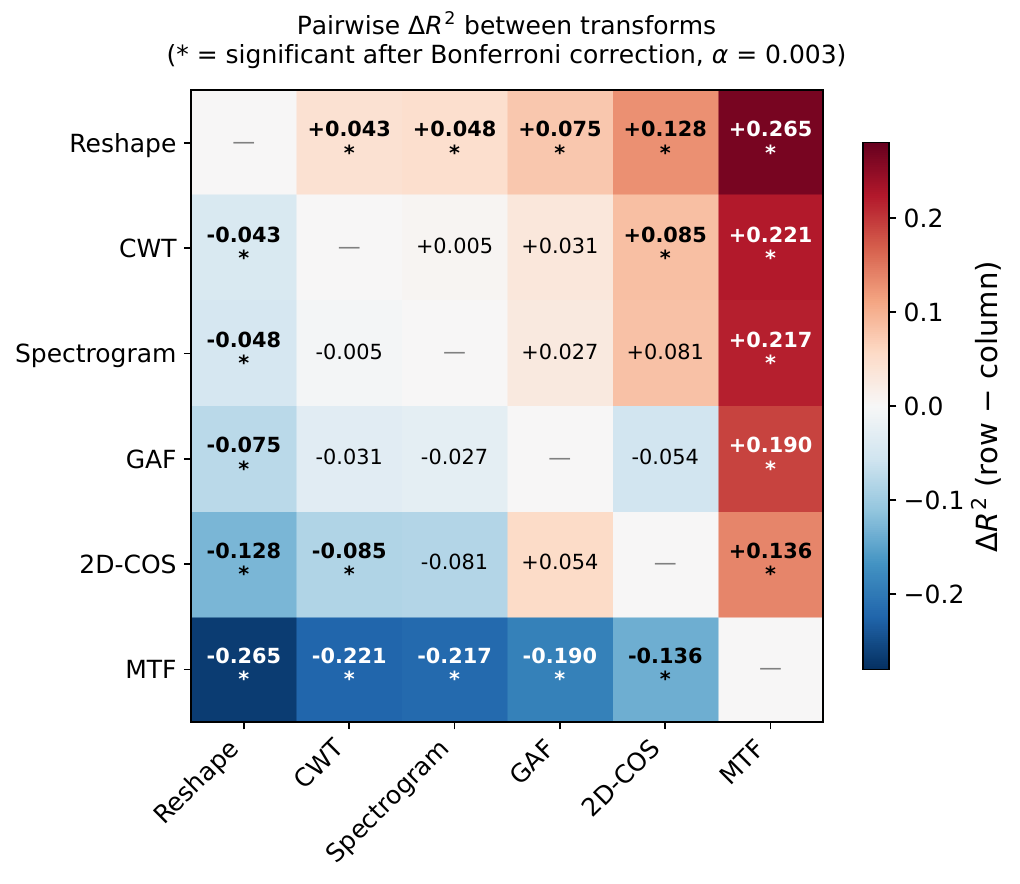}
    \caption{Pairwise differences in mean $R^2$ between the ten 1D-to-2D encodings (the nine transforms and the composite) on the GreenHyperSpectra test set, ordered from best to worst. Each cell gives $\Delta R^2$ = row $-$ column, the difference in mean $R^2$ across the eight traits and three seeds. Blue indicates the row encoding outperforms the column and red the opposite. An asterisk ($^{*}$) marks a pair whose difference is significant (Wilcoxon signed-rank test across the eight traits, $p < 0.05$).}
    \label{fig:pairwise_heatmap}
\end{figure}

\begin{figure}[htbp]
    \centering
    \includegraphics[width=0.85\textwidth]{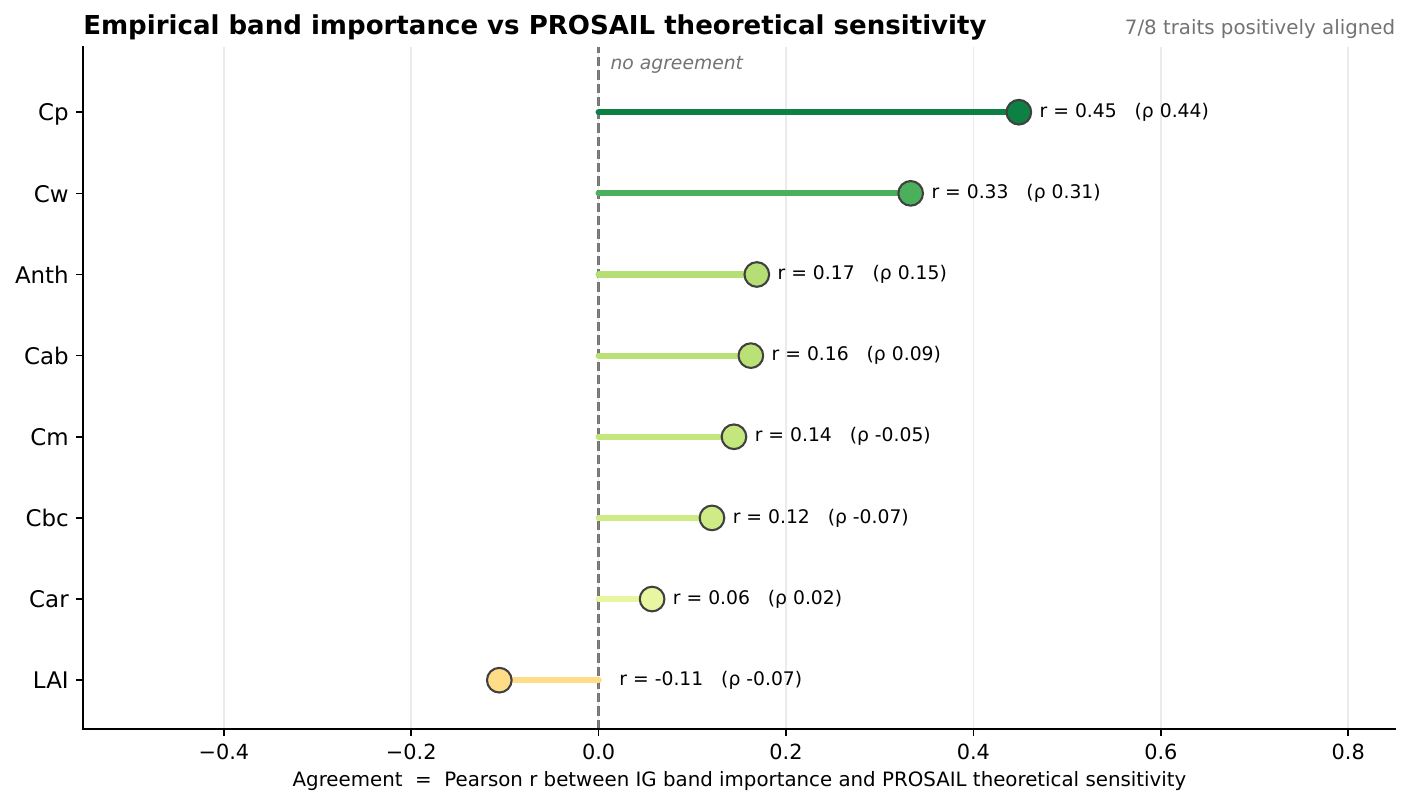}
    \caption{\jl{Per-trait agreement between the Integrated Gradients importance and the PROSAIL theoretical spectral sensitivity, given as the Pearson correlation $r$ and the Spearman correlation $\rho$ across all 1721 bands. Positive values indicate that the model concentrates importance where radiative transfer predicts the constituent acts. Seven of the eight traits are positively aligned.}}
    \label{fig:agreement}
\end{figure}

\begin{figure}[htbp]
    \centering
    \includegraphics[width=\textwidth]{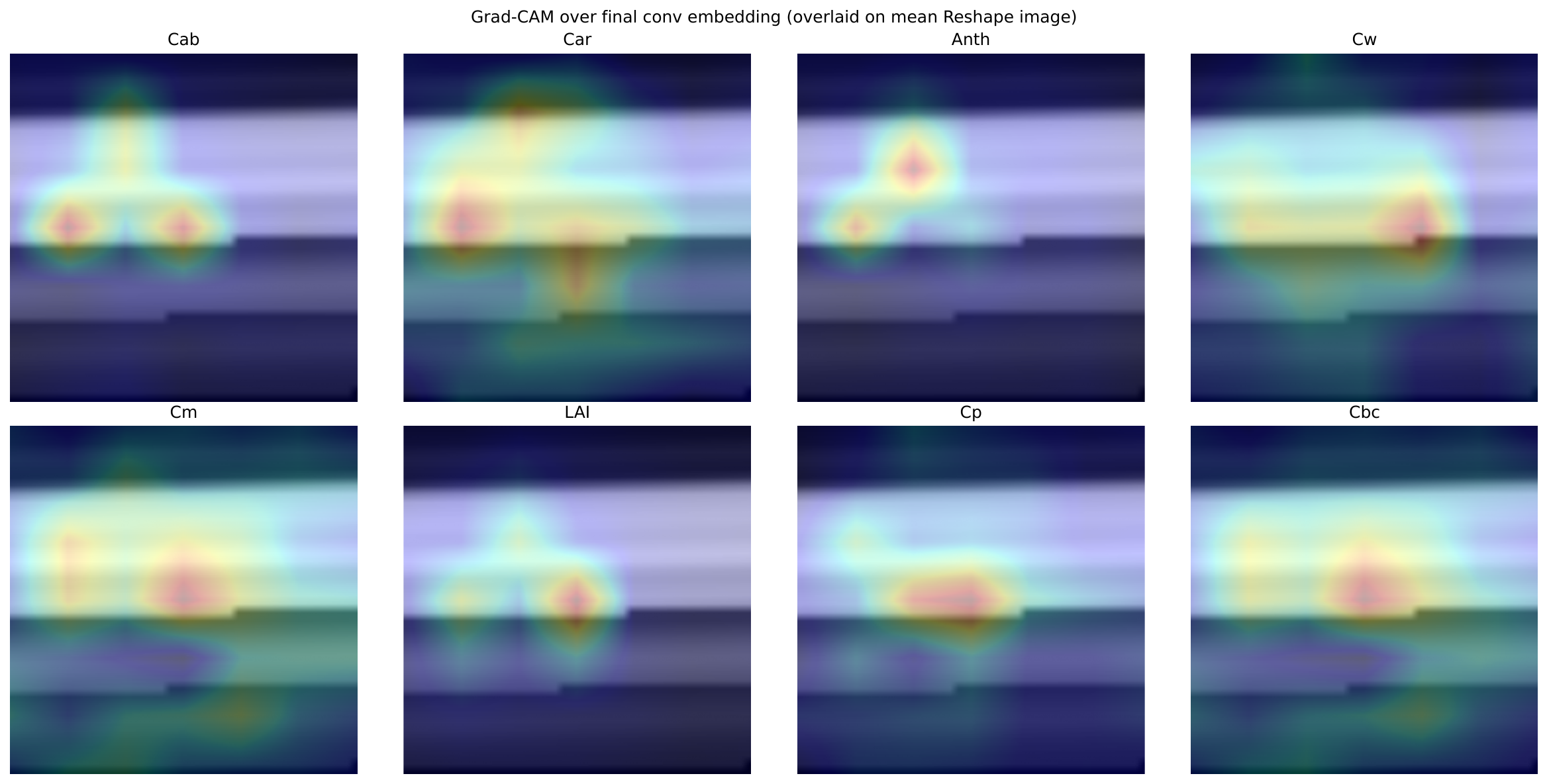}
    \caption{\jl{Grad-CAM relevance maps on the Reshape image for the eight traits. Warmer colors indicate higher relevance for the predicted trait.}}
    \label{fig:gradcam_2d}
\end{figure}

\begin{figure}[htbp]
    \centering
    \includegraphics[width=\textwidth]{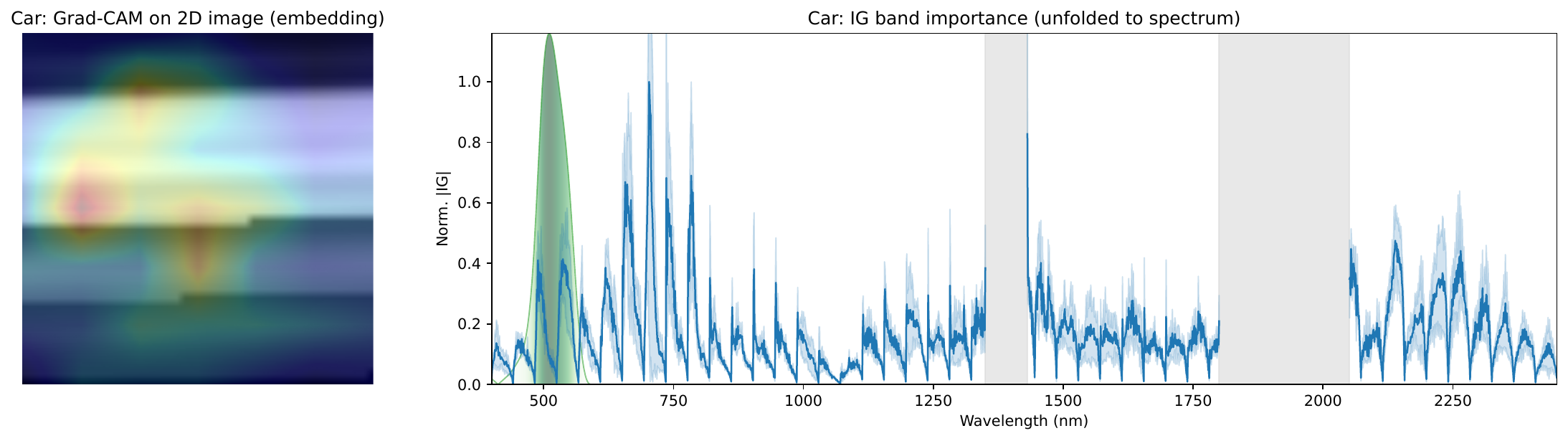}
    \caption{\jl{Linking 2D and 1D attribution for carotenoids (C\textsubscript{ar}).}}
    \label{fig:linking_car}
\end{figure}

\begin{figure}[htbp]
    \centering
    \includegraphics[width=\textwidth]{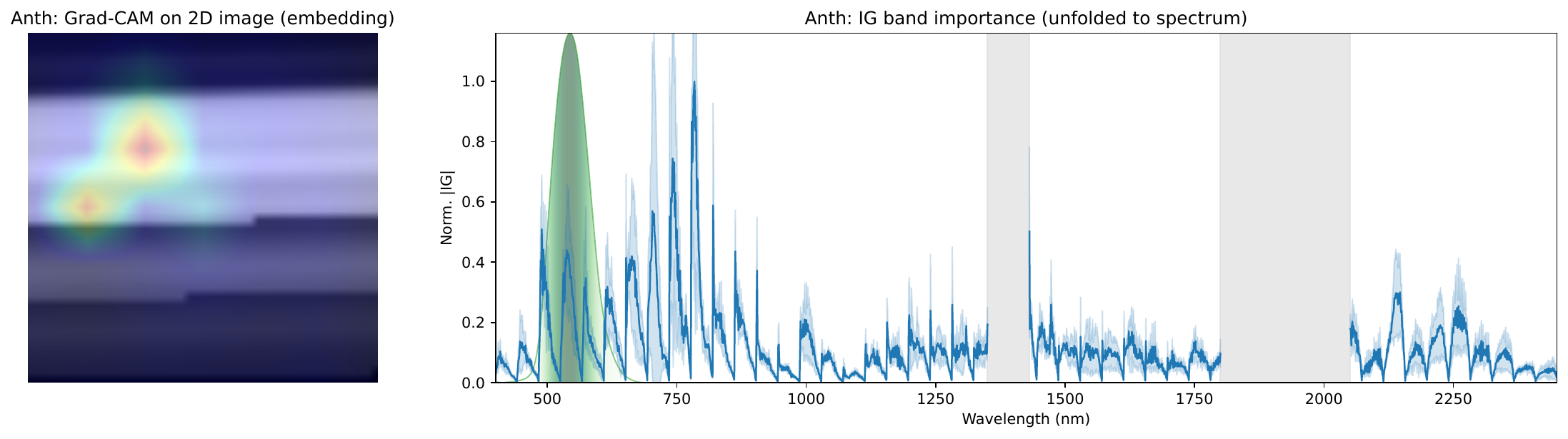}
    \caption{\jl{Linking 2D and 1D attribution for anthocyanins (C\textsubscript{anth}).}}
    \label{fig:linking_anth}
\end{figure}

\begin{figure}[htbp]
    \centering
    \includegraphics[width=\textwidth]{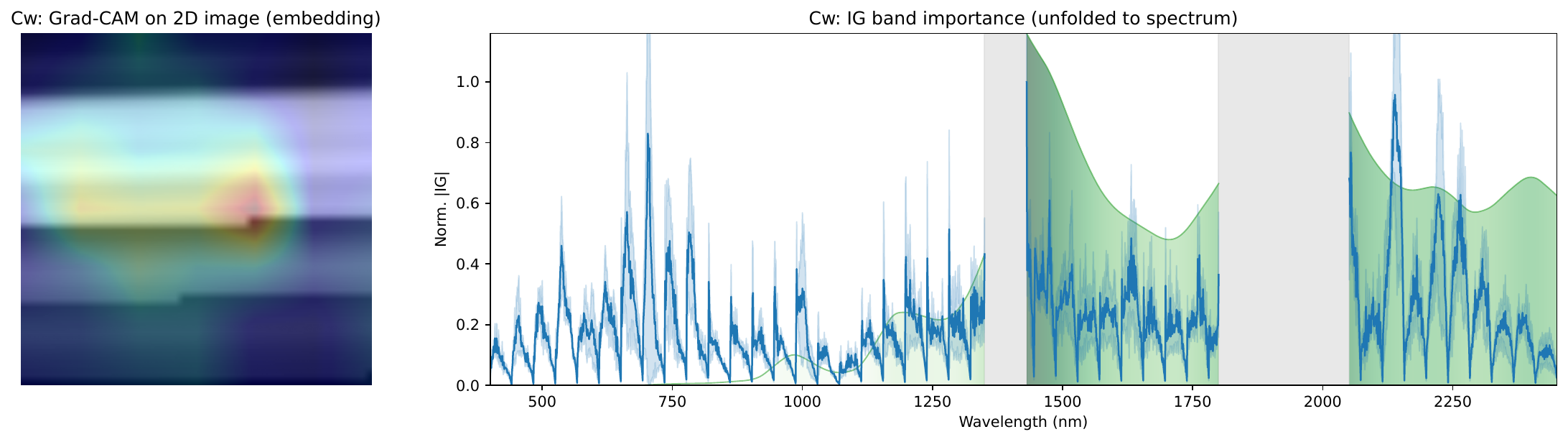}
    \caption{\jl{Linking 2D and 1D attribution for equivalent water thickness (C\textsubscript{w}).}}
    \label{fig:linking_cw}
\end{figure}

\begin{figure}[htbp]
    \centering
    \includegraphics[width=\textwidth]{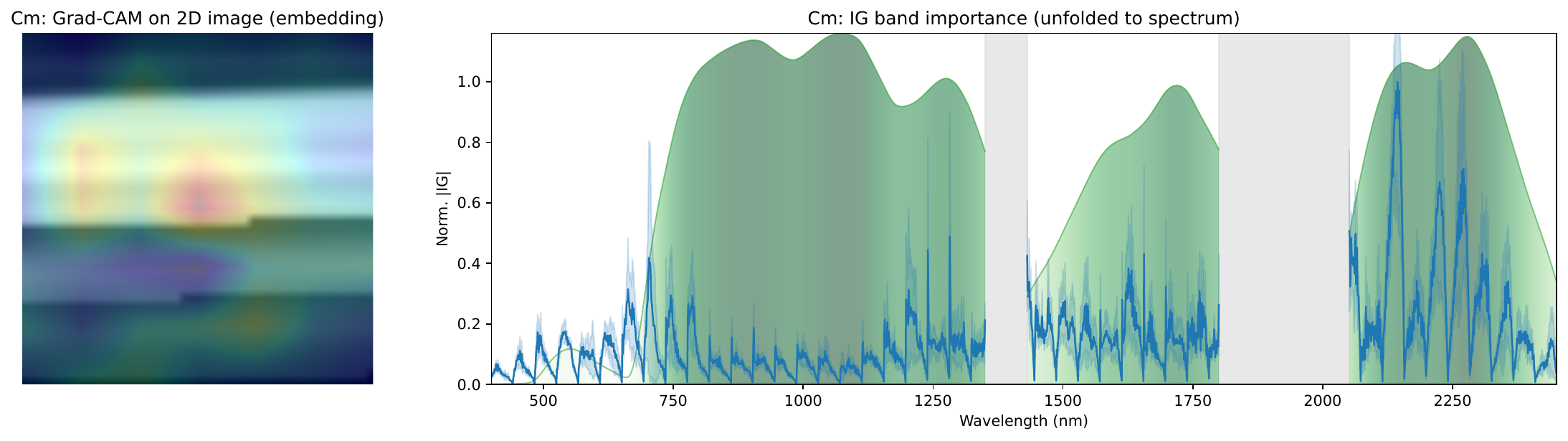}
    \caption{\jl{Linking 2D and 1D attribution for leaf mass per area (C\textsubscript{m}).}}
    \label{fig:linking_cm}
\end{figure}

\begin{figure}[htbp]
    \centering
    \includegraphics[width=\textwidth]{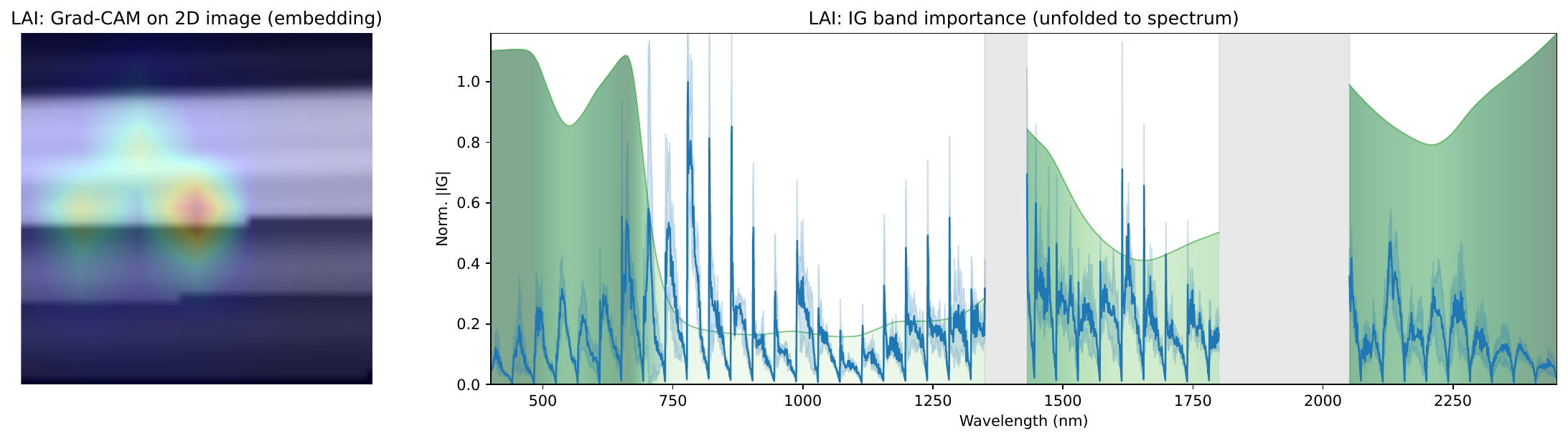}
    \caption{\jl{Linking 2D and 1D attribution for leaf area index (LAI).}}
    \label{fig:linking_lai}
\end{figure}

\begin{figure}[htbp]
    \centering
    \includegraphics[width=\textwidth]{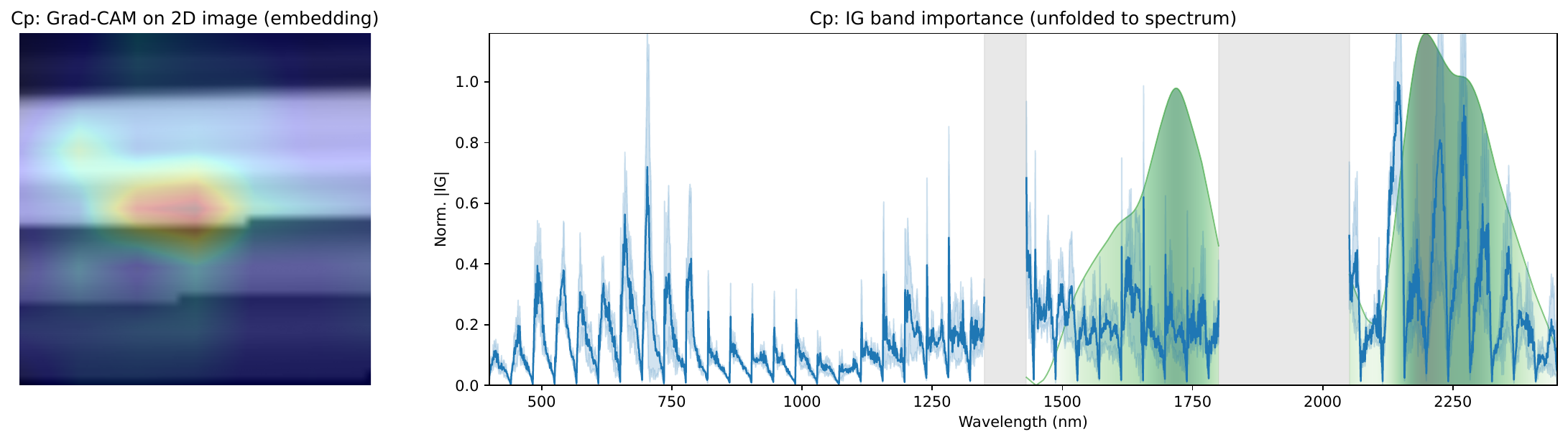}
    \caption{\jl{Linking 2D and 1D attribution for protein content (C\textsubscript{p}).}}
    \label{fig:linking_cp}
\end{figure}

\begin{figure}[htbp]
    \centering
    \includegraphics[width=\textwidth]{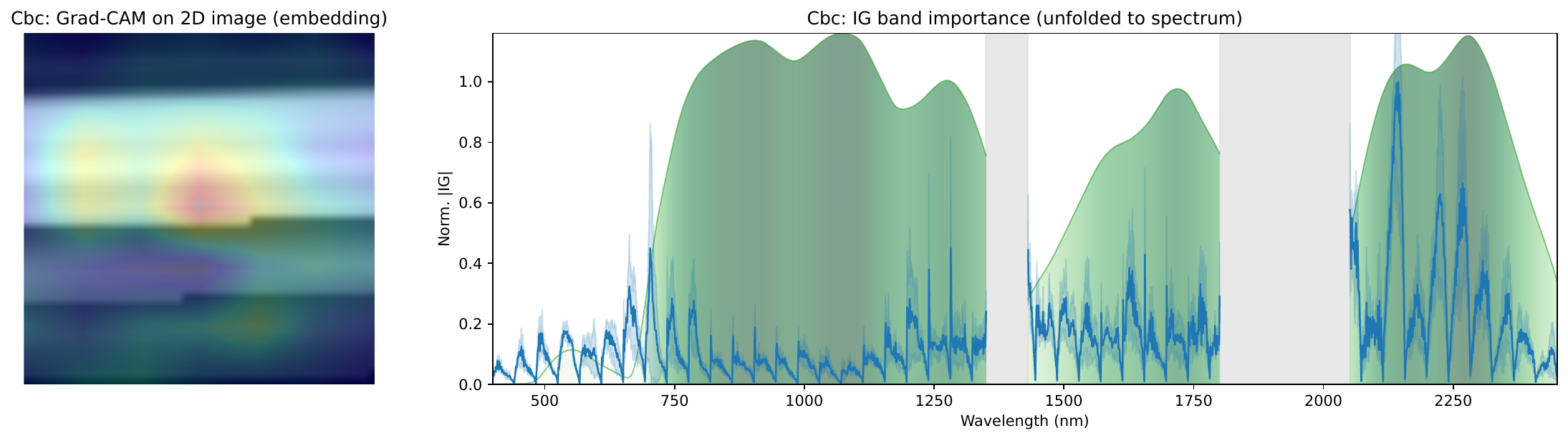}
    \caption{\jl{Linking 2D and 1D attribution for carbon-based constituents (C\textsubscript{bc}).}}
    \label{fig:linking_cbc}
\end{figure}

\jl{Figures~\ref{fig:linking_car} to~\ref{fig:linking_cbc} link the 2D Grad-CAM map and the unfolded 1D importance profile for the remaining seven traits.}

